\documentclass{article}
\usepackage{amssymb}

\usepackage[final]{corl_2025} 

\usepackage{amsmath}
\usepackage{graphicx}
\usepackage{subcaption}
\usepackage{multirow}
\usepackage{tabularx}
\usepackage{booktabs}
\usepackage{colortbl}
\definecolor{RowColor}{rgb}{0.85, 0.85, 0.85}
\usepackage{wrapfig}
\usepackage{enumitem}

\usepackage{cleveref}
\crefname{figure}{Fig.}{Figs.}
\crefname{table}{Tab.}{Tabs.}
\crefname{section}{Sec.}{Secs.}
\crefname{subsection}{Subsec.}{Subsecs.}
\crefname{equation}{Eq.}{Eqs.}
\crefname{theorem}{Thm.}{Thms.}
\crefname{lemma}{Lem.}{Lems.}
\crefname{algorithm}{Alg.}{Algs.}

\title{VLM-AD: End-to-End Autonomous Driving through\\Vision-Language Model Supervision}

\author{
    Yi Xu$^{1,2}$\thanks{Work done at Cruise LLC (GM).} \quad Yuxin Hu$^{3*}$ \quad 
    Zaiwei Zhang$^{4*}$ \quad 
    Gregory P. Meyer$^{4*}$ \quad \\
    \textbf{Siva Karthik Mustikovela$^{1}$ \quad
    Siddhartha Srinivasa$^{5*}$ \quad 
    Eric M. Wolff$^{1}$ \quad 
    Xin Huang$^{6*}$} \\
    $^{1}$ Cruise LLC (GM) \quad
    $^{2}$ Northeastern University \quad
    $^{3}$ OpenAI \\
    $^{4}$ Meta \quad
    $^{5}$ University of Washington \quad
    $^{6}$ Waymo LLC\\
    \texttt{xu.yi@northeastern.edu} \quad
    \texttt{eric.wolff@gm.com} \quad
    \texttt{xhuang@csail.mit.edu}
}

\begin{document}
\maketitle
\begin{abstract}
Human drivers rely on commonsense reasoning to navigate diverse and dynamic real-world scenarios. Existing end-to-end (E2E) autonomous driving (AD) models are typically optimized to mimic driving patterns observed in data, without capturing the underlying reasoning processes.  This limitation constrains their ability to handle challenging driving scenarios. To close this gap, we propose VLM-AD, a method that leverages vision-language models (VLMs) as teachers to enhance training by providing additional supervision that incorporates unstructured reasoning information and structured action labels. Such supervision enhances the model's ability to learn richer feature representations that capture the rationale behind driving patterns. Importantly, our method does not require a VLM during inference, making it practical for real-time deployment. 
When integrated with state-of-the-art methods, VLM-AD achieves significant improvements in planning accuracy and reduced collision rates on the nuScenes dataset. It further improves route completion and driving scores under closed-loop evaluation, demonstrating its effectiveness in long-horizon, interactive driving scenarios and its potential for safe and reliable real-world deployment.
\end{abstract}

\keywords{End-to-End Autonomous Driving, Vision-Language Model, Multimodal Large Language Model, Foundation Models for Driving}

\section{Introduction}
\label{sec:intro}
End-to-end autonomous driving (AD) unifies perception, prediction, and planning into a single framework.
This integration aims to coordinate multiple complex tasks, including detection, tracking, mapping, prediction, and planning. 
Recent approaches~\cite{hu2022st,hu2023planning,jiang2023vad} have tackled these challenges by using sensor data to generate planned ego trajectories with a single, holistic model. 
Although these methods have shown promising results, their performance degrades in challenging, long-tail events~\cite{chib2023recent,chen2024end}.
On the other hand, human drivers often handle such scenarios effectively by reasoning through the driving environment and adapting their actions accordingly. 
This highlights a training gap in current E2E models, which rely solely on the trajectory supervision as sequences of points, lacking the reasoning information necessary for learning rich and robust feature representations to achieve better driving performance.

Manual annotation of reasoning information is often costly, time-consuming, and prone to inconsistent and subjective results, making it difficult to obtain high-quality and scalable annotations. 
Large foundation models offer an alternative by providing their reasoning capabilities for complex tasks such as driving. 
Recent methods~\cite{mao2023gpt,chen2024driving,jin2023adapt,wang2023drivemlm,tian2024drivevlm,xu2024drivegpt4,pan2024vlp,wang2024omnidrive,ding2024hint,hwang2024emma} have directly integrated large foundation models, such as large language models (LLMs)~\cite{devlin2018bert,radford2018improving,touvron2023llama} and vision-language models (VLMs)~\cite{radford2021learning,li2022blip,li2023blip,liu2024visual}, into AD systems to leverage their reasoning capabilities.
However, these 
\begin{wrapfigure}{r}{0.5\textwidth}
    \centering
    \includegraphics[width=\linewidth]{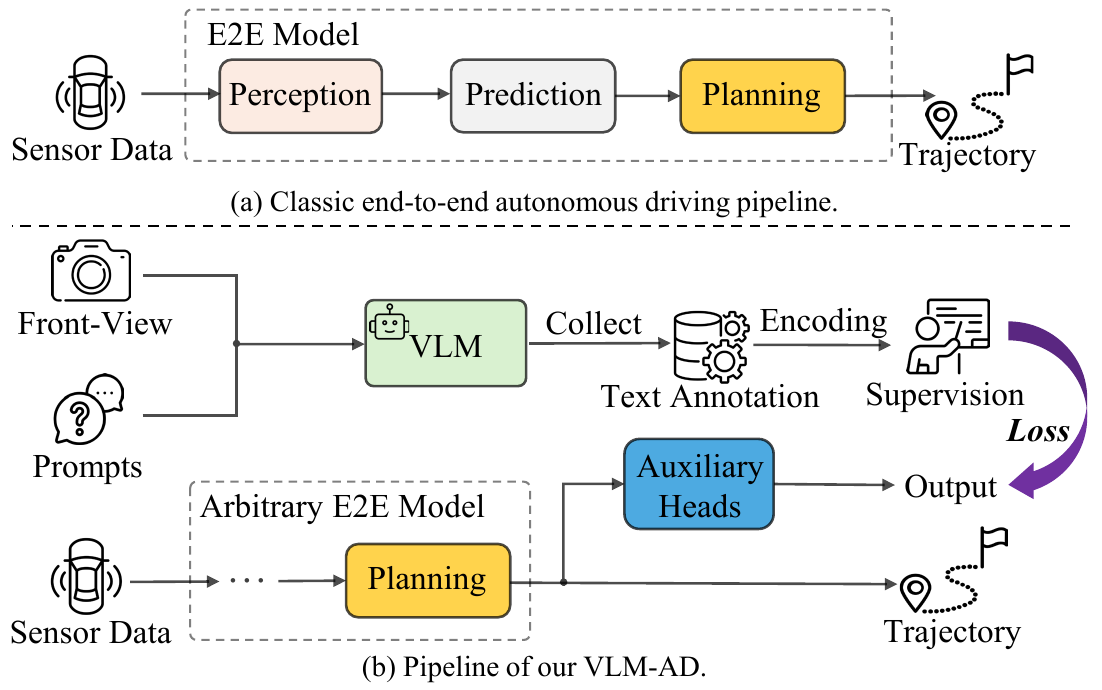}
    \caption{VLM-AD augments an arbitrary end-to-end driving model using auxiliary text prediction tasks during training. These tasks distill driving reasoning knowledge from a VLM to encourage the model to learn richer representations, without fine-tuning a VLM at training time or requiring a VLM at inference time.}
    \vspace{-5mm}
    \label{fig:teaser}
\end{wrapfigure}
methods require extensive fine-tuning to translate language-based outputs into precise numerical results, such as planned trajectories or control signals.
In addition, these methods rely on large foundation models during inference, which significantly increases both training costs and inference time, making these methods impractical for real-world applications.
Given the limitations of manual annotation and the challenges of directly integrating large foundation models into driving systems, we pose the following question: Can large foundation models, such as VLMs, generate reasoning-based text information to enhance autonomous driving models without requiring integration at inference time?

Motivated by this question, we propose VLM-AD, illustrated in \cref{fig:teaser}, a novel method that leverages VLMs as teachers to automatically generate reasoning-based text annotations. 
These annotations then serve as supplementary supervisory signals to train end-to-end pipelines, extending beyond standard trajectory labels. 
Specifically, given a sequence of multi-view images and the future trajectory of the ego vehicle, we project the future trajectory onto the initial front-view image to incorporate critical temporal movement information. 
We then prompt the VLM model with questions regarding the vehicle's current status, intended future actions, and reasoning process to generate both freeform and structured responses, thus infusing critical VLM knowledge into the training pipeline. 

This scalable approach enables us to build a dataset enriched with VLM-generated annotations, effectively addressing the absence of reasoning cues in existing driving datasets. 
We design auxiliary tasks based on these annotations and integrate them seamlessly into existing end-to-end models for joint training. 
These tasks encourage the model to learn richer feature representations for improved driving performance, without requiring VLM involvement at inference time.
Our contributions can be summarized as follows:
\begin{itemize}[leftmargin=*, labelsep=0.5em, itemindent=0pt] 
    \vspace{-0.5mm}
    \item We propose VLM-AD, a simple yet effective approach that distills driving reasoning knowledge from VLMs into end-to-end AD pipelines through a high-quality dataset of reasoning-based behavioral text annotations, generated through carefully crafted prompts directed to VLMs.
    \item We design two plug-and-play auxiliary tasks to supervise existing end-to-end AD pipelines through both unstructured freeform text and structured action labels. 
    These tasks enable effective distillation of VLM knowledge, guiding the model to learn richer feature representations for improved planning performance, without requiring VLM fine-tuning or inference-time usage.
    \item Extensive experiments demonstrate significant improvements in L2 planning accuracy and collision rate over UniAD, VAD, and SparseDrive on the nuScenes dataset, as well as higher route completion and driving scores on the closed-loop CARLA Town05 benchmark.
    \vspace{-0.5mm}
\end{itemize}
\section{Related Work}
\label{sec:related}
\noindent
\textbf{End-to-End Autonomous Driving.}
End-to-end autonomous driving systems jointly train all modules toward a unified goal, resulting in reduced information loss throughout the pipeline.
Unified frameworks such as ST-P3~\cite{hu2022st} and UniAD~\cite{hu2023planning} propose vision-based end-to-end AD systems that unify perception, prediction, and planning.
These models achieve state-of-the-art results on the open-loop nuScenes dataset~\cite{caesar2020nuscenes}.
Following works, such as VAD~\cite{jiang2023vad} and VADv2~\cite{chen2024vadv2}, introduce a vectorized encoding approach for efficient scene representation and extend to closed-loop simulation on CARLA~\cite{dosovitskiy2017carla}.
Recent methods like Ego-MLP~\cite{zhai2023rethinking}, BEV-Planner~\cite{li2024ego}, and PARA-Drive~\cite{weng2024drive} have been developed to explore ego-status and novel design spaces within modular stacks to further enhance driving performance.
In parallel, methods like GenAD~\cite{zheng2024genad}, GraphAD~\cite{zhang2024graphad}, SparseAD~\cite{zhang2024sparsead}, and SparseDrive~\cite{sun2024sparsedrive} continue to push the boundaries of end-to-end driving systems through improved representations and task formulations.
While E2E driving models show promising results in the development of E2E driving methods, they are primarily optimized to mimic driving patterns in the data, without capturing the underlying reasoning processes. This limitation is largely due to the lack of reasoning information in existing datasets. 
Consequently, these methods are unable to acquire deeper reasoning knowledge, which could limit their performance in challenging scenarios.

\noindent
\textbf{Foundation Models for Autonomous Driving.}
Foundation models, including large-language models (LLMs) and vision-language models (VLMs), are being increasingly applied in autonomous driving to leverage their advanced reasoning capabilities. 
GPT-Driver~\cite{mao2023gpt} and Driving-with-LLMs~\cite{chen2024driving} use LLMs to provide action recommendations with explanations, thus enhancing decision transparency. 
A recent approach~\cite{cui2024receive} leverages LLMs to evaluate lane occupancy and safety, enabling more human-like intuitive scene understanding. 
However, LLM-based methods primarily rely on language inputs, which limits their potential to incorporate rich visual features essential for driving.

VLMs address this gap by integrating language and vision for multimodal reasoning, supporting tasks like scene understanding~\cite{qian2024nuscenes,inoue2024nuscenes,choudhary2024talk2bev,sima2023drivelm} and data generation~\cite{jia2023adriver,wang2023drivedreamer,zhao2024drivedreamer}. 
VLMs have also been used for unified navigation and planning~\cite{keysan2023can,tian2024drivevlm,wang2023bevgpt,fu2024drive} as well as end-to-end autonomous driving~\cite{jin2023adapt,wang2023drivemlm,xu2024drivegpt4,pan2024vlp,liu2025vlm}.
However, existing VLM-based methods often require extensive domain-specific fine-tuning, which significantly increases computational cost and inference latency. 
Closely related to our method in end-to-end autonomous driving, VLP~\cite{pan2024vlp} transforms ground-truth trajectory and bounding box labels into text features for contrastive learning, but it does not introduce information beyond existing supervision labels. 
In contrast, our method leverages VLMs to provide additional reasoning information to further enhance driving performance.

\section{Method}
\label{sec:method}

\begin{figure*}[!t]
  \centering
   \includegraphics[width=\linewidth]{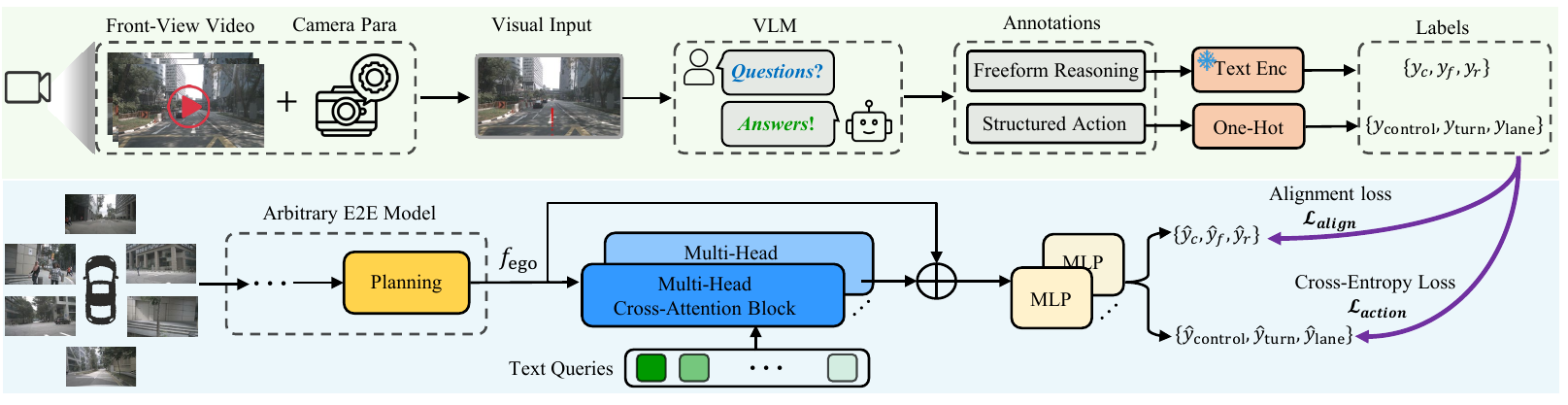}
   \caption{Framework of our proposed VLM-AD. We leverage a VLM as a teacher to generate both freeform reasoning and structured action annotations, which are converted into supervisory signals to enable the model to learn richer representations through auxiliary text alignment and action classification heads. As a result, our method offers better planning results and interpretable action predictions, without requiring a VLM at inference time.}
   \vspace{-4mm}
   \label{fig:fra}
\end{figure*}

\cref{fig:fra} presents an overview of our proposed VLM-AD framework, which consists of two main components. The first is the annotation component, where a VLM generates rich auxiliary information, forming a supplementary dataset for supervision. The second component is our designed auxiliary heads designed to align with this supervision, which can be seamlessly integrated into any E2E model following the planning module.

\subsection{VLM Text Annotation}
\label{sec:vlm_text}




We utilize a VLM as the teacher to enrich the dataset with additional information, leveraging its reasoning capabilities from visual inputs to deepen an E2E model's understanding of driving behaviors. The annotation process can be defined as:
\begin{equation} 
\label{eq:vlm} 
\mathcal{A} = \mathbf{M}(\mathcal{P}, \mathcal{V}),
\end{equation}
where $\mathbf{M}(\cdot)$ represents the VLM model, $\mathcal{P}$ denotes the language prompts, $\mathcal{V}$ is the visual input, and $\mathcal{A}$ is the model's natural language output, serving as annotations for the dataset. 
Our goal is to provide images captured from the ego vehicle's camera, along with specifically crafted prompts, to obtain detailed informative responses from the VLM, leveraging its extensive world knowledge.

In our work, we employ GPT-4o~\cite{achiam2023gpt}, a high-performance VLM trained on internet-scale data, to automatically annotate our dataset. GPT-4o is used to interpret the scenario, generate suitable reasoning-based responses, and identify the actions of the ego vehicle in complex scenarios. 

\noindent
\textbf{Visual Input.}
When determining the visual input, we encounter two challenges. 
The first challenge is selecting the appropriate image(s) from multiple cameras that provide 360-degree coverage around the ego vehicle. 
We explore two approaches: creating a composite large image from all views or using only the front-view image, which typically contains the most relevant information needed for most driving tasks. 
Our annotation results show that both methods yield comparable output quality, so we opt for the front-view image alone to reduce overall complexity.

The second challenge involves integrating temporal information, which is essential for effective planning and decision-making. 
We also consider two approaches. 
One straightforward approach is to input several consecutive frames as a sequence, with prompts that indicate the future timestamps.
However, we observe that VLMs struggle with temporal continuity and often confuse the ego vehicle's identity, likely due to limitations in temporal grounding~\cite{qiu2023large,kesen2023vilma}. 
Instead, we project the ego vehicle's future trajectory onto a single front-view image, leveraging the camera's intrinsic and extrinsic parameters along with sensor specifications. 
We specify in the prompts that the projected trajectory reflects the vehicle's future path. 
This cost-effective design allows the VLM to interpret temporal information more reliably than using an image sequence.

\noindent
\textbf{Freeform Reasoning Annotation.}
As a key input to the VLM, a well-designed question is essential for enhancing reasoning capabilities~\cite{wei2022chain} and improving the explainability of the VLM's responses. 
In our approach, we focus on the planning task, by designing prompts specifically to obtain reasoning from the VLM. 
We create two types of questions, beginning with open-ended questions intended to generate free-form, unstructured responses that contain rich, high-dimensional language information. 
We refer to these responses as unstructured reasoning annotations.

To maximize the reasoning capabilities of the VLM, we provide detailed context descriptions as preliminary instructions before posing specific questions, defined as follows:
\begin{itemize}[leftmargin=*, labelsep=0.5em, itemindent=0pt]
    \vspace{-0.5mm}
    \item $C_{1}$: \textit{This is the front-view image of the ego vehicle. The red line indicates the future trajectory, no line suggests stopping or slowing down. When explaining the reasoning, please focus on the camera image and the surrounding context rather than referencing the plotted trajectory.}
    \vspace{-0.5mm}
    \item $Q_{1-1}$: \textit{Please describe the ego vehicle's current actions.} 
     \vspace{-0.5mm}
    \item $Q_{1-2}$: \textit{Please predict the ego vehicle's future actions.}
     \vspace{-0.5mm}
    \item $Q_{1-3}$: \textit{Please explain the reasoning of current and future action.}
     \vspace{-0.5mm}
\end{itemize}

The full input prompt is defined as $\mathcal{P}_{1} = [C_{1}, Q_{1}]$, where $Q_{1} = \{Q_{1-1}, Q_{1-2}, Q_{1-3}\}$ represents the set of questions. These open-ended questions yield free-form text annotations describing the ego vehicle's current status, intended future actions, and the reasoning underlying the VLM's knowledge.

\noindent
\textbf{Structured Action Annotation.}
To examine the flexibility of our method, we define a second type of question in a structured format. Specifically, we create three distinct action sets and prompt the VLM to select answers from these predefined options. This allows us to obtain a single action annotation for each question. Specifically, the context and questions are defined as follows:
\begin{itemize}[leftmargin=*, labelsep=0.5em, itemindent=0pt]
    \vspace{-0.5mm}
    \item $C_{2}$: \textit{This is the front-view image of the ego vehicle. The red line indicates the future trajectory, no line suggests stopping or slowing down.}
    \vspace{-0.5mm}
    \item $Q_{2-1}$: \textit{Please describe the ego vehicle's action from the control action list: \{go straight, move slowly, stop, reverse\}.}
    \vspace{-0.5mm}
    \item $Q_{2-2}$: \textit{Please describe the ego vehicle's action from the turn action list: \{turn left, turn right, turn around, none\}.}
    \vspace{-0.5mm}
    \item $Q_{2-3}$: \textit{Please describe the ego vehicle's action from the lane action list: \{change lane to the left, change lane to the right, merge into the left lane, merge into the right lane, none\}.}
    \vspace{-0.5mm}
\end{itemize}

The complete input prompt is defined as $\mathcal{P}_{2} = [C_{2}, Q_{2}]$, where $Q_{2} = \{Q_{2-1}, Q_{2-2}, Q_{2-3}\}$ represents the set of structured action questions. In this way, we can obtain three specific actions from the VLM. Compared to freeform text annotations, one major benefit of structured annotations is that they can be used to supervise an E2E driving model to predict human-interpretable actions, as demonstrated in the experimental results of~\cref{sec:exp}.

\subsection{Auxiliary Heads}
Typically, data-driven end-to-end autonomous driving methods~\cite{hu2023planning,jiang2023vad} focus on summarizing a learnable ego feature $f_{\text{ego}}$ to produce planning results, which is essential for generating reliable and accurate planning trajectories. 
This learnable ego feature aggregates all relevant information about the ego vehicle from upstream modules through different networks.
In our approach, we develop auxiliary heads that use this ego feature as input, enabling the model to distill knowledge from the VLM's responses.

\noindent
\textbf{Annotation Encoding.}
Using the $Q_{1}$ questions, we obtain three text responses, denoted as $\mathcal{A}_{1} = \{\mathcal{A}_{c}, \mathcal{A}_{f}, \mathcal{A}_{r}\}$, which represent descriptions of the current action, future action prediction, and reasoning, respectively.
Using the $Q_{2}$ questions, we obtain three actions from predefined sets, denoted as $\mathcal{A}_{2} = \{\mathcal{A}_{\text{control}}, \mathcal{A}_{\text{turn}}, \mathcal{A}_{\text{lane}}\}$, corresponding to the control action, turn action, and lane action.
To convert these annotations into supervisory signals, we apply two distinct approaches to generate two corresponding types of labels, effectively integrating them into end-to-end autonomous driving pipelines as supervision.

For the freeform text annotations from $Q_{1}$, we utilize an off-the-shelf language model, such as CLIP~\cite{radford2021learning}, to convert the text into feature representations. For the structured answers, each action is encoded as a one-hot label. Formally:
\begin{equation}\label{eq:pro}
y_{1} = \text{CLIP}(\mathcal{A}_{1}), \quad y_{2} = \text{One-Hot}(\mathcal{A}_{2}),
\end{equation}
where $y_{1}$ and $y_{2}$ each have three components: $y_{1} = \{y_{c}, y_{f}, y_{r}\}$ and $y_{2} = \{y_{\text{control}}, y_{\text{turn}}, y_{\text{lane}}\}$. 
Here, $y_{c}$, $y_{f}$, and $y_{r}$ are feature vectors of size $C$, where $C$ is the dimension of text embedding, while $y_{\text{control}}$, $y_{\text{turn}}$, and $y_{\text{lane}}$ are three one-hot action labels with size $N$, where $N_{\text{control}}=4$, $N_{\text{turn}}=4$ and $N_{\text{lane}}=5$, respectively.

\noindent
\textbf{Text Feature Alignment.}
Using the three text features $y_{1} = \{y_{c}, y_{f}, y_{r}\}$ as supervision, we develop a feature alignment head that takes the ego feature $f_{\text{ego}}$ as input. 
This setup resembles knowledge distillation, 
where the alignment head is trained to match the text features provided by the VLM.

In this head, we initialize three learnable text queries, $q_{1} = \{q_{c}, q_{f}, q_{r}\}$. 
Each query interacts with the ego feature $f_{\text{ego}}$ via a multi-head cross-attention (MHCA) block, where the text query acts as the attention query $q$, and the ego feature serves as both the key $k$ and value $v$, producing updated text queries. 
These updated queries are then concatenated with the ego feature to form the feature representation for this text head, which is subsequently processed through an MLP layer to produce the final feature alignment output. This process is formulated as:
\begin{equation} \label{eq:head_1}
    q'_{1} = \text{MHCA}(q, k, v), \quad
    q=q_{1}, k=v=f_{\text{ego}}, \quad
    \hat{f}_{1} = \text{MLP}(q'_{1}  \oplus  f_{\text{ego}}),
\end{equation}
where $\oplus$ denotes concatenation, and $\hat{f}_{1}=\{\hat{f}_{c}, \hat{f}_{f}, \hat{f}_{r}\}$ represents three output features to be aligned with the corresponding VLM text features. 
Note that we use three independent MHCA blocks, one for each component, enabling each text query to focus on specific aspects of the ego feature that can be represented in text form.

Inspired by the knowledge distillation approach DINO~\cite{caron2021emerging} that controls the smoothness and sharpness of features, we adopt a similar strategy to normalize the text and output features with different temperature parameters, producing feature distributions rather than raw feature values as follows:
\begin{equation} \label{eq:dino}
P(y_{1}) = \frac{\exp(y_{1} / \tau_{t})}{\sum_{k=1}^{C} \exp(y_{1}^{(k)} / \tau_{t})}, \quad
P(\hat{f}_{1}) = \frac{\exp(\hat{f}_{1} / \tau_{s})}{\sum_{k=1}^{C} \exp(\hat{f}_{1}^{(k)} / \tau_{s})},
\end{equation}
where $\tau_{t}$ and $\tau_{s}$ are temperature parameters that control the sharpness of these distributions. 
This adjustment enables better alignment between the output features and supervisory labels, enhancing alignment quality for knowledge distillation. 
Note that we do not apply the centering operation, as we consider the supervision to be ground truth.

\noindent
\textbf{Structured Action Classification.}
We obtain the structured action labels $y_{2} = \{y_{\text{control}}, y_{\text{turn}}, y_{\text{lane}}\}$ from the VLM using question $Q_{2}$.
We then construct another action classification head that takes the ego feature $f_{\text{ego}}$ as input.
Similar to the previous feature alignment stage, we initialize three learnable action queries, $q_{\text{control}}$, $q_{\text{turn}}$, and $q_{\text{lane}}$, which interact with $f_{\text{ego}}$ through three MHCA blocks. 
In this setup, each action query serves as the attention query $q$, while the ego feature acts as the key $k$ and value $v$, producing updated action queries.
We then concatenate updated queries with the ego feature to create the representation for the action classification head, passing it through an MLP layer followed by a Softmax to generate the action predictions. This process is formulated as:
\begin{equation}
    q'_{2} = \text{MHCA}(q,k,v), \quad
    q=q_{2}, k=v=f_{\text{ego}}, \quad
    \hat{f}_{2} = \text{Softmax}\left(\text{MLP}(q'_{2}  \oplus f_{\text{ego}})\right),
\end{equation}
where $\hat{f}_{2}=\{\hat{f}_{\text{control}}, \hat{f}_{\text{turn}}, \hat{f}_{\text{lane}}\}$ represents the predicted control action, turn action, and lane action. 
We use independent MHCA blocks for each action query to produce distinct action labels.

\subsection{Auxiliary Loss}
We define two parallel auxiliary tasks following the planning module to enable the model to distill knowledge from the VLM, and the overall training loss is a weighted sum of two components:
\begin{equation} 
\mathcal{L} = \lambda_{1}\mathcal{L}_{\text{align}} + \lambda_{2}\mathcal{L}_{\text{action}},  
\end{equation}
where each component corresponds to a distinct auxiliary text head to provide supervision:
\begin{equation}
    \label{eq:loss}
    \begin{aligned}
    \mathcal{L}_{\text{align}}= &
    - P(y_{c})\log(P(\hat{f}_{c})) 
    - P(y_{f})\log(P(\hat{f}_{f})) - P(y_{r})\log(P(\hat{f}_{r}))\\
    \mathcal{L}_{\text{action}}= &
    -\sum_{i=1}^{N_{\text{control}}} y_{\text{control}}^{i} \log(\hat{f}_{\text{control}}^{i}) -\sum_{i=1}^{N_{\text{turn}}} y_{\text{turn}}^{i} \log(\hat{f}_{\text{turn}}^{i}) -\sum_{i=1}^{N_{\text{lane}}} y_{\text{lane}}^{i} \log(\hat{f}_{\text{lane}}^{i})
    \end{aligned}
    .
\end{equation}

For feature alignment, we use cross-entropy loss to align the supervisory and output features, capturing the critical information conveyed by the text. For the action classification task, we also apply cross-entropy loss to ensure accurate classification.


\section{Experiments}
\label{sec:exp}
\subsection{Open-Loop Settings}
\noindent
\textbf{Baselines.}
Our proposed method is a general framework compatible with various end-to-end autonomous driving methods. 
We validate its effectiveness by applying it to three widely recognized open-source methods, UniAD~\cite{hu2023planning}, VAD~\cite{jiang2023vad}, and SparseDrive~\cite{sun2024sparsedrive}.
Additionally, we compare it with VLP~\cite{pan2024vlp}, which projects ego vehicle ground-truth labels into text feature space via CLIP~\cite{radford2021learning} for contrastive learning.


\noindent
\textbf{Dataset.}
We use the nuScenes dataset~\cite{caesar2020nuscenes} for open-loop planning evaluation. nuScenes is a large-scale autonomous driving dataset featuring 1000 scenes, each with a duration of approximately 20 seconds and annotated at 2Hz. The dataset includes detailed annotations, making it a popular benchmark for end-to-end autonomous driving research. 

\noindent
\textbf{Metrics.}
We focus on the planning task and use standard metrics such as L2 displacement error and collision rate to evaluate performance for open-loop evaluation.

\subsection{Closed-Loop Settings}
We further evaluate our proposed method in a closed-loop setting. Following the protocol established by VAD~\cite{jiang2023vad}, we conduct simulations using the CARLA~\cite{dosovitskiy2017carla} simulator and the Town05~\cite{prakash2021multi} benchmark. To ensure a fair comparison, we integrate our method into the VAD framework and adopt the same evaluation definitions for closed-loop evaluation.

\noindent
\textbf{Metrics.}
We use the standard Route Completion (RC) and Driving Score (DS) metrics for closed-loop performance evaluation.

\noindent
\subsection{Implementation Details}
We use official codes of the UniAD, VAD, and SparseDrive, adhering to the hyperparameters specified in their official implementations. For our proposed VLM-AD, we define two auxiliary task heads, each containing an MHCA block with 8 heads and 3 cross-attention layers, and we set 3 text queries for each of $Q_{1}$ and $Q_{2}$. During training, we set the temperature parameters $\tau_{s}=0.1$ and $\tau_{t}=0.04$ to control the sharpness of the features, and we set $\lambda_{1}=1$ and $\lambda_{2}=0.1$ to balance $\mathcal{L}_{align}$ and $\mathcal{L}_{action}$. All models are trained on 8 NVIDIA H100 GPUs using the PyTorch framework~\cite{paszke2019pytorch}. Complete implementation details, annotation quality analysis, and additional experiments, including visualizations, are provided in the appendix.


\subsection{Main Results}
\begin{table*}[t]
\centering
\caption{Planning results of our VLM-AD method and baselines. The best results are shown in bold and the second best are underlined.  VLM-AD consistently outperforms the baselines, with the reasoning-focused $Q_{1}$ contributing the most significant improvements.}
\scalebox{0.85}{
\begin{tabular}{c|l|cc|cccc|cccc|c}
\toprule
\multicolumn{1}{l|}{\multirow{2}{*}{ID}} &  \multicolumn{1}{l|}{\multirow{2}{*}{Method}} & \multicolumn{1}{l}{\multirow{2}{*}{$Q_1$}} & \multicolumn{1}{l|}{\multirow{2}{*}{$Q_2$}} &\multicolumn{4}{c|}{L2 (m) $\downarrow$} & \multicolumn{4}{c|}{Collision Rate (\%) $\downarrow$} & \multicolumn{1}{l}{\multirow{2}{*}{Ckpt. Source}}\\
& & & & 1s & 2s & 3s & Avg. & 1s & 2s & 3s & Avg. & \\
\midrule

0 
& UniAD~\cite{hu2023planning} 
& &
& 0.48 
& 0.96 
& 1.65 
& 1.03
& \underline{0.05} 
& 0.17 
& 0.71 
& 0.31 
& Official \\

1
& UniAD*
& &
& 0.46 & 0.96 & 1.67 & 1.03 & 0.11 & 0.22 & 0.74 & 0.36 
& Reproduced \\

2 
& VLP~\cite{pan2024vlp}
& &
& 0.43 & 0.86 & 1.47 & 0.92 & \textbf{0.03} & 0.15& \underline{0.48} & \underline{0.22} 
& Official \\


\rowcolor{RowColor} 
3
& VLM-AD
& \checkmark & 
& \underline{0.40} & \underline{0.83} & \underline{1.44} & \underline{0.89} & \underline{0.05} & \textbf{0.11} & 0.56 & 0.24 & - \\

\rowcolor{RowColor} 
4
& VLM-AD
&  & \checkmark
& 0.41 & 0.87 & 1.46 & 0.91 & 0.06 & \underline{0.13} & 0.68 & 0.29 & - 
\\

\rowcolor{RowColor} 
5
& VLM-AD
& \checkmark & \checkmark
& \textbf{0.39} & \textbf{0.82} & \textbf{1.43} & \textbf{0.88} 
& \underline{0.05} & \textbf{0.11} & \textbf{0.43} & \textbf{0.19} & - 
\\

\cmidrule(lr){1-13}

6
& VAD-Base~\cite{jiang2023vad}
& &
& 0.41 & 0.70 & 1.06 & 0.72 & \textbf{0.04} & 0.43 & 1.15 & 0.54
& Official
\\

7
& VAD-Base$^{\#}$
& &
& 0.33 & 0.59 & 0.94 & 0.62 & 0.19 & 0.30 & 0.53 & 0.34
& Reproduced \\

8
& VLP~\cite{pan2024vlp}
& &
& \underline{0.26} & \underline{0.47} & 0.78 & 0.50
& \underline{0.12} & \underline{0.17} & \underline{0.42} & \textbf{0.23}
& Official \\


\rowcolor{RowColor} 
9
& VLM-AD
& \checkmark &
& \textbf{0.24} & 0.48 & \underline{0.76} & \underline{0.49} & \underline{0.12} & \textbf{0.16} & \textbf{0.41} & \textbf{0.23} 
& -
\\

\rowcolor{RowColor} 
10
& VLM-AD
& & \checkmark
& 0.30 & 0.50 & 0.82 & 0.54 & 0.16 & 0.24 & 0.43 & \underline{0.28}
& -
\\

\rowcolor{RowColor} 
11
& VLM-AD
& \checkmark & \checkmark 
& \textbf{0.24} & \textbf{0.46} & \textbf{0.75} & \textbf{0.48} & \underline{0.12} & \underline{0.17} & \textbf{0.41} & \textbf{0.23}
& -
\\

\cmidrule(lr){1-13}

12
& VAD-Tiny~\cite{jiang2023vad}
& &
& 0.46 & 0.76 & 1.12 & 0.78 & 0.21 & 0.35 & 0.58 & 0.38
& Official
\\

13
& VAD-Tiny$^{\#}$
& &
& 0.35 & 0.62 & 0.96 & 0.64 & 0.12 & 0.19 & 0.44 & 0.25
& Reproduced \\


\rowcolor{RowColor} 
14
& VLM-AD
& \checkmark &
& \textbf{0.30} & \textbf{0.54} & \underline{0.82} & \textbf{0.55} & \textbf{0.08} & \textbf{0.15} & \textbf{0.38} & \textbf{0.20} & -
\\

\rowcolor{RowColor} 
15
& VLM-AD
& & \checkmark 
& \underline{0.31} & \underline{0.55} & 0.88 & \underline{0.58} & \underline{0.10} &\underline{0.18} & \underline{0.41} & 0.23 & -

\\
\rowcolor{RowColor} 
16
& VLM-AD
& \checkmark & \checkmark 
& \textbf{0.30} & \textbf{0.54} & \textbf{0.80} & \textbf{0.55} & 0.11 & \textbf{0.15} & \textbf{0.38} & \underline{0.21} & -
\\

\cmidrule(lr){1-13}

17 
& SparseDrive-B~\cite{sun2024sparsedrive} 
& &
& 0.29 & 0.55 & 0.91 & 0.58 & \textbf{0.01} & \textbf{0.02} & 0.13 & \underline{0.06}
& Official \\

\rowcolor{RowColor} 
18 
& VLM-AD
& \checkmark &
& \underline{0.25} & \underline{0.50} & \underline{0.86} & \underline{0.54}
& \textbf{0.01} & \textbf{0.02} & \textbf{0.12} & \textbf{0.05} & -\\

\rowcolor{RowColor} 
19 
& VLM-AD
& & \checkmark 
& 0.27 & 0.52 & 0.87 & 0.55
& \textbf{0.01} & \textbf{0.02} & \underline{0.13} & \textbf{0.05} & -\\

\rowcolor{RowColor} 
20 
& VLM-AD
& \checkmark & \checkmark 
&\textbf{0.24} & \textbf{0.49} & \textbf{0.84} & \textbf{0.52}
& \textbf{0.01} & \textbf{0.02} & \textbf{0.12} & \textbf{0.05} & -\\
\bottomrule
\end{tabular}
}
\label{tab:main}
\end{table*}

\noindent
\textbf{Open-Loop Planning Results.}
\cref{tab:main} presents the results of applying our proposed VLM-AD to three baselines, UniAD, VAD, and SparseDrive, as well as comparisons with VLP.
Comparing methods ID 0 and 1, we achieve nearly identical planning results using the authors' officially trained checkpoints. 
For methods IDs 6 and 7, and IDs 12 and 13, we observe some discrepancies between our reproduced results and the reported values, which we attribute to a correction in image configuration in the official codebase\footnote{\url{https://github.com/hustvl/VAD/issues/9}}.
From the first section of the table, we observe that VLM-AD significantly outperforms UniAD on both average L2 planning error and average collision rate by introducing $Q_{1}$ and $Q_{2}$. It also surpasses a state-of-the-art baseline VLP in both metrics. Regarding VAD, our VLM-AD consistently outperforms both VAD-Base and VAD-Tiny, particularly on the L2 planning error metric, and it achieves superior performance compared to the VLP in VAD-Base.
For SparseDrive-B, we observe that integrating our VLM-AD leads to improvements in L2 planning error over the baseline model, while maintaining comparable or slightly better collision rates. This is largely due to the collision-aware rescore module in the baseline, which already effectively reduces collision rates. Importantly, the integration of VLM-AD does not compromise this strength, further demonstrating the robustness of our approach.
These results demonstrate the effectiveness and advantages of our VLM-AD approach. Additionally, $Q_{1}$ yields better results than $Q_{2}$, verifying the value of supervising the driving model through rich reasoning information.

\begin{wraptable}{r}{0.5\textwidth}
\centering
\vspace{-4mm}
\caption{Closed-loop evaluation results of our VLM-AD method and baselines. The best is in bold and the second best is underlined. $^{\dag}$: LiDAR-based method.}
\scalebox{0.9}{
\begin{tabular}{l|cc|cc}
\toprule
\multicolumn{1}{l|}{\multirow{2}{*}{Method}} & \multicolumn{2}{c}{Town05 Short} &  \multicolumn{2}{c}{Town05 Long}\\
& DS $\uparrow$ & RC $\uparrow$ & DS $\uparrow$ & RC $\uparrow$ \\
\midrule
CILRS~\cite{codevilla2019exploring} 
& 7.47 & 13.40 & 3.68 & 7.19 \\

LBC~\cite{chen2020learning}
& 30.97 & 55.01 & 7.05 & 32.09 \\

Transfuser$^{\dag}$~\cite{prakash2021multi}
& 54.52 & 78.41 & \underline{33.15} & 56.36 \\

ST-P3~\cite{hu2022st} 
& 55.14 & 86.74 & 11.45 & \underline{83.15} \\

VAD-Base~\cite{jiang2023vad}
& \underline{64.29} & \underline{87.26} & 30.31 & 75.20 \\

\rowcolor{RowColor}
+VLM-AD
& \textbf{67.78} & \textbf{88.56} & \textbf{35.25} & \textbf{84.14}\\

\bottomrule
\end{tabular}
}
\label{tab:closedloop}
\end{wraptable}
\noindent
\textbf{Closed-Loop Planning Results.}
\cref{tab:closedloop} presents the closed-loop results on the Town05 benchmark for both short and long routes. Our proposed method, VLM-AD, achieves the best performance across all metrics, including Driving Score (DS) and Route Completion (RC). 
It surpasses VAD-Base and ST-P3, two strong vision-based baselines, with noticeable gains particularly in DS on the long route. These results demonstrate the effectiveness and robustness of VLM-AD in realistic closed-loop driving scenarios, highlighting its ability to reason and act reliably under long-horizon tasks.

\subsection{Ablation Study}
\label{sec:ablation_study}
\begin{table*}[t]
\centering
\caption{Ablation study on the contributions of $Q_{1-1}$, $Q_{1-2}$ and $Q_{1-3}$ to the UniAD baseline. The best is in bold and the second best is underlined.}
\scalebox{0.94}{
\begin{tabular}{l|cccc|cccc|cccc}
\toprule
\multicolumn{1}{l|}{\multirow{2}{*}{Method}} & \multicolumn{1}{l}{\multirow{2}{*}{$Q_{1-1}$}} & \multicolumn{1}{l}{\multirow{2}{*}{$Q_{1-2}$}} & \multicolumn{1}{l}{\multirow{2}{*}{$Q_{1-3}$}} &
\multicolumn{1}{l|}{\multirow{2}{*}{$Q_{2}$}}&\multicolumn{4}{c|}{L2 (m) $\downarrow$} & \multicolumn{4}{c}{Collision Rate (\%) $\downarrow$} \\
& & & & & 1s & 2s & 3s & Avg. & 1s & 2s & 3s & Avg. \\
\midrule

UniAD
& & & &
& 0.48 & 0.96 & 1.65 & 1.03 & \underline{0.05} & 0.17 & 0.71 & 0.31 \\

VLM-AD
& \checkmark & \checkmark & & 
& \underline{0.41} & 0.87 & 1.53 & 0.94 & 0.08 & 0.17 & 0.61 & 0.29
\\

VLM-AD
& \checkmark & & \checkmark & 
& \underline{0.41} & \underline{0.84} & \textbf{1.44} & \underline{0.90} & \underline{0.05} & \textbf{0.11} & \textbf{0.48} & \textbf{0.21}
\\

VLM-AD
& & \checkmark & \checkmark & 
& 0.42 & 0.87 & \underline{1.49} & 0.93 & \textbf{0.03} & \underline{0.14} & \underline{0.51} & \underline{0.23}
\\

VLM-AD
& \checkmark & \checkmark & \checkmark & 
& \textbf{0.40} & \textbf{0.83} & \textbf{1.44} & \textbf{0.89} & \underline{0.05} & \textbf{0.11} & 0.56 & 0.24
\\

\bottomrule
\end{tabular}
}
\label{tab:ablation}
\end{table*}
\noindent
We further analyze the contributions of each sub-question, $Q_{1-1}$, $Q_{1-2}$, and $Q_{1-3}$, within $Q_{1}$. 
Each sub-question provides specific text information related to the ego vehicle’s current status, predicted future actions, and reasoning.
\cref{tab:ablation} presents an ablation study of these three sub-questions. The results indicate that each sub-question positively impacts the overall performance, demonstrating that our designed questions provide valuable information for the planning task. Notably, the reasoning feature contributes the most to reducing the L2 planning error, underscoring its importance in enhancing driving performance. We defer additional ablation studies to Sec.~\ref{sec:appendix_ablation_study}.

\section{Conclusion}
\label{sec:con}
In this work, we presented VLM-AD, a novel approach to enhancing end-to-end autonomous driving models by leveraging vision-language models (VLMs) as auxiliary teachers. By integrating VLM-based annotations through targeted questions that include unstructured reasoning text and structured action labels, we enriched the training process with additional reasoning and action supervision. Our method significantly improves planning accuracy and collision rates in the nuScenes dataset, while also achieving higher route completion and driving scores under closed-loop evaluation.
Importantly, it does not require VLMs at inference time, making it plug-and-play for real-world deployment without additional inference overhead. 

\section*{Limitations}
\label{sec:lim}
One limitation of our work is its focus solely on the planning task, with questions centered on behavior and action. In future work, our approach could be extended to generate relevant annotations for perception or prediction, potentially benefiting the entire autonomous driving pipeline. 
Additionally, we plan to apply our method to more challenging datasets and design a broader set of questions to extract diverse and useful knowledge from VLMs.


	




\bibliography{ref}  

\begin{thebibliography}{73}
\providecommand{\natexlab}[1]{#1}
\providecommand{\url}[1]{\texttt{#1}}
\expandafter\ifx\csname urlstyle\endcsname\relax
  \providecommand{\doi}[1]{doi: #1}\else
  \providecommand{\doi}{doi: \begingroup \urlstyle{rm}\Url}\fi

\bibitem[Hu et~al.(2022)Hu, Chen, Wu, Li, Yan, and Tao]{hu2022st}
S.~Hu, L.~Chen, P.~Wu, H.~Li, J.~Yan, and D.~Tao.
\newblock St-p3: End-to-end vision-based autonomous driving via spatial-temporal feature learning.
\newblock In \emph{ECCV}, pages 533--549, 2022.

\bibitem[Hu et~al.(2023)Hu, Yang, Chen, Li, Sima, Zhu, Chai, Du, Lin, Wang, et~al.]{hu2023planning}
Y.~Hu, J.~Yang, L.~Chen, K.~Li, C.~Sima, X.~Zhu, S.~Chai, S.~Du, T.~Lin, W.~Wang, et~al.
\newblock Planning-oriented autonomous driving.
\newblock In \emph{CVPR}, pages 17853--17862, 2023.

\bibitem[Jiang et~al.(2023)Jiang, Chen, Xu, Liao, Chen, Zhou, Zhang, Liu, Huang, and Wang]{jiang2023vad}
B.~Jiang, S.~Chen, Q.~Xu, B.~Liao, J.~Chen, H.~Zhou, Q.~Zhang, W.~Liu, C.~Huang, and X.~Wang.
\newblock Vad: Vectorized scene representation for efficient autonomous driving.
\newblock In \emph{ICCV}, pages 8340--8350, 2023.

\bibitem[Chib and Singh(2023)]{chib2023recent}
P.~S. Chib and P.~Singh.
\newblock Recent advancements in end-to-end autonomous driving using deep learning: A survey.
\newblock \emph{IEEE TIV}, 2023.

\bibitem[Chen et~al.(2024)Chen, Wu, Chitta, Jaeger, Geiger, and Li]{chen2024end}
L.~Chen, P.~Wu, K.~Chitta, B.~Jaeger, A.~Geiger, and H.~Li.
\newblock End-to-end autonomous driving: Challenges and frontiers.
\newblock \emph{IEEE TPAMI}, 2024.

\bibitem[Mao et~al.(2023)Mao, Qian, Ye, Zhao, and Wang]{mao2023gpt}
J.~Mao, Y.~Qian, J.~Ye, H.~Zhao, and Y.~Wang.
\newblock Gpt-driver: Learning to drive with gpt.
\newblock In \emph{NeurIPS Workshop}, 2023.

\bibitem[Chen et~al.(2024)Chen, Sinavski, H{\"u}nermann, Karnsund, Willmott, Birch, Maund, and Shotton]{chen2024driving}
L.~Chen, O.~Sinavski, J.~H{\"u}nermann, A.~Karnsund, A.~J. Willmott, D.~Birch, D.~Maund, and J.~Shotton.
\newblock Driving with llms: Fusing object-level vector modality for explainable autonomous driving.
\newblock In \emph{ICRA}, pages 14093--14100, 2024.

\bibitem[Jin et~al.(2023)Jin, Liu, Zheng, Li, Zhao, Zhang, Zheng, Zhou, and Liu]{jin2023adapt}
B.~Jin, X.~Liu, Y.~Zheng, P.~Li, H.~Zhao, T.~Zhang, Y.~Zheng, G.~Zhou, and J.~Liu.
\newblock Adapt: Action-aware driving caption transformer.
\newblock In \emph{ICRA}, pages 7554--7561, 2023.

\bibitem[Wang et~al.(2023)Wang, Xie, Hu, Zou, Fan, Tong, Wen, Wu, Deng, Li, et~al.]{wang2023drivemlm}
W.~Wang, J.~Xie, C.~Hu, H.~Zou, J.~Fan, W.~Tong, Y.~Wen, S.~Wu, H.~Deng, Z.~Li, et~al.
\newblock Drivemlm: Aligning multi-modal large language models with behavioral planning states for autonomous driving.
\newblock \emph{arXiv preprint arXiv:2312.09245}, 2023.

\bibitem[Tian et~al.(2024)Tian, Gu, Li, Liu, Hu, Wang, Zhan, Jia, Lang, and Zhao]{tian2024drivevlm}
X.~Tian, J.~Gu, B.~Li, Y.~Liu, C.~Hu, Y.~Wang, K.~Zhan, P.~Jia, X.~Lang, and H.~Zhao.
\newblock Drivevlm: The convergence of autonomous driving and large vision-language models.
\newblock \emph{arXiv preprint arXiv:2402.12289}, 2024.

\bibitem[Xu et~al.(2024)Xu, Zhang, Xie, Zhao, Guo, Wong, Li, and Zhao]{xu2024drivegpt4}
Z.~Xu, Y.~Zhang, E.~Xie, Z.~Zhao, Y.~Guo, K.-Y.~K. Wong, Z.~Li, and H.~Zhao.
\newblock Drivegpt4: Interpretable end-to-end autonomous driving via large language model.
\newblock \emph{IEEE RA-L}, 2024.

\bibitem[Pan et~al.(2024)Pan, Yaman, Nesti, Mallik, Allievi, Velipasalar, and Ren]{pan2024vlp}
C.~Pan, B.~Yaman, T.~Nesti, A.~Mallik, A.~G. Allievi, S.~Velipasalar, and L.~Ren.
\newblock Vlp: Vision language planning for autonomous driving.
\newblock In \emph{CVPR}, pages 14760--14769, 2024.

\bibitem[Wang et~al.(2024)Wang, Yu, Jiang, Lan, Shi, Chang, Kautz, Li, and Alvarez]{wang2024omnidrive}
S.~Wang, Z.~Yu, X.~Jiang, S.~Lan, M.~Shi, N.~Chang, J.~Kautz, Y.~Li, and J.~M. Alvarez.
\newblock Omnidrive: A holistic llm-agent framework for autonomous driving with 3d perception, reasoning and planning.
\newblock \emph{arXiv preprint arXiv:2405.01533}, 2024.

\bibitem[Ding et~al.(2024)Ding, Chen, Su, Gao, Jin, Sima, Zhang, Li, Barsch, Li, et~al.]{ding2024hint}
K.~Ding, B.~Chen, Y.~Su, H.-a. Gao, B.~Jin, C.~Sima, W.~Zhang, X.~Li, P.~Barsch, H.~Li, et~al.
\newblock Hint-ad: Holistically aligned interpretability in end-to-end autonomous driving.
\newblock \emph{arXiv preprint arXiv:2409.06702}, 2024.

\bibitem[Hwang et~al.(2024)Hwang, Xu, Lin, Hung, Ji, Choi, Huang, He, Covington, Sapp, et~al.]{hwang2024emma}
J.-J. Hwang, R.~Xu, H.~Lin, W.-C. Hung, J.~Ji, K.~Choi, D.~Huang, T.~He, P.~Covington, B.~Sapp, et~al.
\newblock Emma: End-to-end multimodal model for autonomous driving.
\newblock \emph{arXiv preprint arXiv:2410.23262}, 2024.

\bibitem[Devlin(2018)]{devlin2018bert}
J.~Devlin.
\newblock Bert: Pre-training of deep bidirectional transformers for language understanding.
\newblock \emph{arXiv preprint arXiv:1810.04805}, 2018.

\bibitem[Radford(2018)]{radford2018improving}
A.~Radford.
\newblock Improving language understanding by generative pre-training.
\newblock 2018.

\bibitem[Touvron et~al.(2023)Touvron, Lavril, Izacard, Martinet, Lachaux, Lacroix, Rozi{\`e}re, Goyal, Hambro, Azhar, et~al.]{touvron2023llama}
H.~Touvron, T.~Lavril, G.~Izacard, X.~Martinet, M.-A. Lachaux, T.~Lacroix, B.~Rozi{\`e}re, N.~Goyal, E.~Hambro, F.~Azhar, et~al.
\newblock Llama: Open and efficient foundation language models.
\newblock \emph{arXiv preprint arXiv:2302.13971}, 2023.

\bibitem[Radford et~al.(2021)Radford, Kim, Hallacy, Ramesh, Goh, Agarwal, Sastry, Askell, Mishkin, Clark, et~al.]{radford2021learning}
A.~Radford, J.~W. Kim, C.~Hallacy, A.~Ramesh, G.~Goh, S.~Agarwal, G.~Sastry, A.~Askell, P.~Mishkin, J.~Clark, et~al.
\newblock Learning transferable visual models from natural language supervision.
\newblock In \emph{ICML}, pages 8748--8763, 2021.

\bibitem[Li et~al.(2022)Li, Li, Xiong, and Hoi]{li2022blip}
J.~Li, D.~Li, C.~Xiong, and S.~Hoi.
\newblock Blip: Bootstrapping language-image pre-training for unified vision-language understanding and generation.
\newblock In \emph{ICML}, pages 12888--12900, 2022.

\bibitem[Li et~al.(2023)Li, Li, Savarese, and Hoi]{li2023blip}
J.~Li, D.~Li, S.~Savarese, and S.~Hoi.
\newblock Blip-2: Bootstrapping language-image pre-training with frozen image encoders and large language models.
\newblock In \emph{ICML}, pages 19730--19742, 2023.

\bibitem[Liu et~al.(2024)Liu, Li, Wu, and Lee]{liu2024visual}
H.~Liu, C.~Li, Q.~Wu, and Y.~J. Lee.
\newblock Visual instruction tuning.
\newblock In \emph{Advances in neural information processing systems}, pages 34892--34916, 2024.

\bibitem[Caesar et~al.(2020)Caesar, Bankiti, Lang, Vora, Liong, Xu, Krishnan, Pan, Baldan, and Beijbom]{caesar2020nuscenes}
H.~Caesar, V.~Bankiti, A.~H. Lang, S.~Vora, V.~E. Liong, Q.~Xu, A.~Krishnan, Y.~Pan, G.~Baldan, and O.~Beijbom.
\newblock nuscenes: A multimodal dataset for autonomous driving.
\newblock In \emph{CVPR}, pages 11621--11631, 2020.

\bibitem[Chen et~al.(2024)Chen, Jiang, Gao, Liao, Xu, Zhang, Huang, Liu, and Wang]{chen2024vadv2}
S.~Chen, B.~Jiang, H.~Gao, B.~Liao, Q.~Xu, Q.~Zhang, C.~Huang, W.~Liu, and X.~Wang.
\newblock Vadv2: End-to-end vectorized autonomous driving via probabilistic planning.
\newblock \emph{arXiv preprint arXiv:2402.13243}, 2024.

\bibitem[Dosovitskiy et~al.(2017)Dosovitskiy, Ros, Codevilla, Lopez, and Koltun]{dosovitskiy2017carla}
A.~Dosovitskiy, G.~Ros, F.~Codevilla, A.~Lopez, and V.~Koltun.
\newblock Carla: An open urban driving simulator.
\newblock In \emph{CoRL}, pages 1--16, 2017.

\bibitem[Zhai et~al.(2023)Zhai, Feng, Du, Mao, Liu, Tan, Zhang, Ye, and Wang]{zhai2023rethinking}
J.-T. Zhai, Z.~Feng, J.~Du, Y.~Mao, J.-J. Liu, Z.~Tan, Y.~Zhang, X.~Ye, and J.~Wang.
\newblock Rethinking the open-loop evaluation of end-to-end autonomous driving in nuscenes.
\newblock \emph{arXiv preprint arXiv:2305.10430}, 2023.

\bibitem[Li et~al.(2024)Li, Yu, Lan, Li, Kautz, Lu, and Alvarez]{li2024ego}
Z.~Li, Z.~Yu, S.~Lan, J.~Li, J.~Kautz, T.~Lu, and J.~M. Alvarez.
\newblock Is ego status all you need for open-loop end-to-end autonomous driving?
\newblock In \emph{CVPR}, pages 14864--14873, 2024.

\bibitem[Weng et~al.(2024)Weng, Ivanovic, Wang, Wang, and Pavone]{weng2024drive}
X.~Weng, B.~Ivanovic, Y.~Wang, Y.~Wang, and M.~Pavone.
\newblock Para-drive: Parallelized architecture for real-time autonomous driving.
\newblock In \emph{CVPR}, pages 15449--15458, 2024.

\bibitem[Zheng et~al.(2024)Zheng, Song, Guo, Zhang, and Chen]{zheng2024genad}
W.~Zheng, R.~Song, X.~Guo, C.~Zhang, and L.~Chen.
\newblock Genad: Generative end-to-end autonomous driving.
\newblock In \emph{European Conference on Computer Vision}, pages 87--104, 2024.

\bibitem[Zhang et~al.(2024{\natexlab{a}})Zhang, Qian, Li, Pan, Chen, Liang, Zhang, Zhang, Li, Fu, et~al.]{zhang2024graphad}
Y.~Zhang, D.~Qian, D.~Li, Y.~Pan, Y.~Chen, Z.~Liang, Z.~Zhang, S.~Zhang, H.~Li, M.~Fu, et~al.
\newblock Graphad: Interaction scene graph for end-to-end autonomous driving.
\newblock \emph{arXiv preprint arXiv:2403.19098}, 2024{\natexlab{a}}.

\bibitem[Zhang et~al.(2024{\natexlab{b}})Zhang, Wang, Zhu, Zhao, Chen, Zhang, Gong, Zhou, Zhang, Wang, et~al.]{zhang2024sparsead}
D.~Zhang, G.~Wang, R.~Zhu, J.~Zhao, X.~Chen, S.~Zhang, J.~Gong, Q.~Zhou, W.~Zhang, N.~Wang, et~al.
\newblock Sparsead: Sparse query-centric paradigm for efficient end-to-end autonomous driving.
\newblock \emph{arXiv preprint arXiv:2404.06892}, 2024{\natexlab{b}}.

\bibitem[Sun et~al.(2025)Sun, Lin, Shi, Zhang, Wu, and Zheng]{sun2024sparsedrive}
W.~Sun, X.~Lin, Y.~Shi, C.~Zhang, H.~Wu, and S.~Zheng.
\newblock {SparseDrive}: End-to-end autonomous driving via sparse scene representation.
\newblock In \emph{ICRA}, 2025.

\bibitem[Cui et~al.(2024)Cui, Ma, Cao, Ye, and Wang]{cui2024receive}
C.~Cui, Y.~Ma, X.~Cao, W.~Ye, and Z.~Wang.
\newblock Receive, reason, and react: Drive as you say, with large language models in autonomous vehicles.
\newblock \emph{IEEE ITSM}, 2024.

\bibitem[Qian et~al.(2024)Qian, Chen, Zhuo, Jiao, and Jiang]{qian2024nuscenes}
T.~Qian, J.~Chen, L.~Zhuo, Y.~Jiao, and Y.-G. Jiang.
\newblock Nuscenes-qa: A multi-modal visual question answering benchmark for autonomous driving scenario.
\newblock In \emph{AAAI}, pages 4542--4550, 2024.

\bibitem[Inoue et~al.(2024)Inoue, Yada, Tanahashi, and Yamaguchi]{inoue2024nuscenes}
Y.~Inoue, Y.~Yada, K.~Tanahashi, and Y.~Yamaguchi.
\newblock Nuscenes-mqa: Integrated evaluation of captions and qa for autonomous driving datasets using markup annotations.
\newblock In \emph{WACV}, pages 930--938, 2024.

\bibitem[Choudhary et~al.(2024)Choudhary, Dewangan, Chandhok, Priyadarshan, Jain, Singh, Srivastava, Jatavallabhula, and Krishna]{choudhary2024talk2bev}
T.~Choudhary, V.~Dewangan, S.~Chandhok, S.~Priyadarshan, A.~Jain, A.~K. Singh, S.~Srivastava, K.~M. Jatavallabhula, and K.~M. Krishna.
\newblock Talk2bev: Language-enhanced bird’s-eye view maps for autonomous driving.
\newblock In \emph{ICRA}, pages 16345--16352, 2024.

\bibitem[Sima et~al.(2024)Sima, Renz, Chitta, Chen, Zhang, Xie, Luo, Geiger, and Li]{sima2023drivelm}
C.~Sima, K.~Renz, K.~Chitta, L.~Chen, H.~Zhang, C.~Xie, P.~Luo, A.~Geiger, and H.~Li.
\newblock Drivelm: Driving with graph visual question answering.
\newblock In \emph{ECCV}, 2024.

\bibitem[Jia et~al.(2023)Jia, Mao, Liu, Zhao, Wen, Zhang, Zhang, and Wang]{jia2023adriver}
F.~Jia, W.~Mao, Y.~Liu, Y.~Zhao, Y.~Wen, C.~Zhang, X.~Zhang, and T.~Wang.
\newblock Adriver-i: A general world model for autonomous driving.
\newblock \emph{arXiv preprint arXiv:2311.13549}, 2023.

\bibitem[Wang et~al.(2023)Wang, Zhu, Huang, Chen, Zhu, and Lu]{wang2023drivedreamer}
X.~Wang, Z.~Zhu, G.~Huang, X.~Chen, J.~Zhu, and J.~Lu.
\newblock Drivedreamer: Towards real-world-driven world models for autonomous driving.
\newblock \emph{arXiv preprint arXiv:2309.09777}, 2023.

\bibitem[Zhao et~al.(2024)Zhao, Wang, Zhu, Chen, Huang, Bao, and Wang]{zhao2024drivedreamer}
G.~Zhao, X.~Wang, Z.~Zhu, X.~Chen, G.~Huang, X.~Bao, and X.~Wang.
\newblock Drivedreamer-2: Llm-enhanced world models for diverse driving video generation.
\newblock \emph{arXiv preprint arXiv:2403.06845}, 2024.

\bibitem[Keysan et~al.(2023)Keysan, Look, Kosman, G{\"u}rsun, Wagner, Yao, and Rakitsch]{keysan2023can}
A.~Keysan, A.~Look, E.~Kosman, G.~G{\"u}rsun, J.~Wagner, Y.~Yao, and B.~Rakitsch.
\newblock Can you text what is happening? integrating pre-trained language encoders into trajectory prediction models for autonomous driving.
\newblock \emph{arXiv preprint arXiv:2309.05282}, 2023.

\bibitem[Wang et~al.(2023)Wang, Zhu, Lu, Zhong, Chen, Shen, Wang, and Wang]{wang2023bevgpt}
P.~Wang, M.~Zhu, H.~Lu, H.~Zhong, X.~Chen, S.~Shen, X.~Wang, and Y.~Wang.
\newblock Bevgpt: Generative pre-trained large model for autonomous driving prediction, decision-making, and planning.
\newblock \emph{arXiv preprint arXiv:2310.10357}, 2023.

\bibitem[Fu et~al.(2024)Fu, Li, Wen, Dou, Cai, Shi, and Qiao]{fu2024drive}
D.~Fu, X.~Li, L.~Wen, M.~Dou, P.~Cai, B.~Shi, and Y.~Qiao.
\newblock Drive like a human: Rethinking autonomous driving with large language models.
\newblock In \emph{WACV}, pages 910--919, 2024.

\bibitem[Liu et~al.(2025)Liu, Liu, Liu, Liu, Ni, and Ma]{liu2025vlm}
P.~Liu, H.~Liu, H.~Liu, X.~Liu, J.~Ni, and J.~Ma.
\newblock Vlm-e2e: Enhancing end-to-end autonomous driving with multimodal driver attention fusion.
\newblock \emph{arXiv preprint arXiv:2502.18042}, 2025.

\bibitem[Achiam et~al.(2023)Achiam, Adler, Agarwal, Ahmad, Akkaya, Aleman, Almeida, Altenschmidt, Altman, Anadkat, et~al.]{achiam2023gpt}
J.~Achiam, S.~Adler, S.~Agarwal, L.~Ahmad, I.~Akkaya, F.~L. Aleman, D.~Almeida, J.~Altenschmidt, S.~Altman, S.~Anadkat, et~al.
\newblock Gpt-4 technical report.
\newblock \emph{arXiv preprint arXiv:2303.08774}, 2023.

\bibitem[Qiu et~al.(2023)Qiu, Zhao, Ziser, Korhonen, Ponti, and Cohen]{qiu2023large}
Y.~Qiu, Z.~Zhao, Y.~Ziser, A.~Korhonen, E.~M. Ponti, and S.~B. Cohen.
\newblock Are large language models temporally grounded?
\newblock \emph{arXiv preprint arXiv:2311.08398}, 2023.

\bibitem[Kesen et~al.(2023)Kesen, Pedrotti, Dogan, Cafagna, Acikgoz, Parcalabescu, Calixto, Frank, Gatt, Erdem, et~al.]{kesen2023vilma}
I.~Kesen, A.~Pedrotti, M.~Dogan, M.~Cafagna, E.~C. Acikgoz, L.~Parcalabescu, I.~Calixto, A.~Frank, A.~Gatt, A.~Erdem, et~al.
\newblock Vilma: A zero-shot benchmark for linguistic and temporal grounding in video-language models.
\newblock \emph{arXiv preprint arXiv:2311.07022}, 2023.

\bibitem[Wei et~al.(2022)Wei, Wang, Schuurmans, Bosma, Xia, Chi, Le, Zhou, et~al.]{wei2022chain}
J.~Wei, X.~Wang, D.~Schuurmans, M.~Bosma, F.~Xia, E.~Chi, Q.~V. Le, D.~Zhou, et~al.
\newblock Chain-of-thought prompting elicits reasoning in large language models.
\newblock In \emph{NeurIPS}, pages 24824--24837, 2022.

\bibitem[car(2021)]{caron2021emerging}
Emerging properties in self-supervised vision transformers.
\newblock In \emph{ICCV}, pages 9650--9660, 2021.

\bibitem[Prakash et~al.(2021)Prakash, Chitta, and Geiger]{prakash2021multi}
A.~Prakash, K.~Chitta, and A.~Geiger.
\newblock Multi-modal fusion transformer for end-to-end autonomous driving.
\newblock In \emph{CVPR}, pages 7077--7087, 2021.

\bibitem[Paszke et~al.(2019)Paszke, Gross, Massa, Lerer, Bradbury, Chanan, Killeen, Lin, Gimelshein, Antiga, et~al.]{paszke2019pytorch}
A.~Paszke, S.~Gross, F.~Massa, A.~Lerer, J.~Bradbury, G.~Chanan, T.~Killeen, Z.~Lin, N.~Gimelshein, L.~Antiga, et~al.
\newblock Pytorch: {An} imperative style, high-performance deep learning library.
\newblock In \emph{NeurIPS}, 2019.

\bibitem[Codevilla et~al.(2019)Codevilla, Santana, L{\'o}pez, and Gaidon]{codevilla2019exploring}
F.~Codevilla, E.~Santana, A.~M. L{\'o}pez, and A.~Gaidon.
\newblock Exploring the limitations of behavior cloning for autonomous driving.
\newblock In \emph{ICCV}, pages 9329--9338, 2019.

\bibitem[Chen et~al.(2020)Chen, Zhou, Koltun, and Kr{\"a}henb{\"u}hl]{chen2020learning}
D.~Chen, B.~Zhou, V.~Koltun, and P.~Kr{\"a}henb{\"u}hl.
\newblock Learning by cheating.
\newblock In \emph{CoRL}, pages 66--75, 2020.

\bibitem[Chitta et~al.(2022)Chitta, Prakash, Jaeger, Yu, Renz, and Geiger]{chitta2022transfuser}
K.~Chitta, A.~Prakash, B.~Jaeger, Z.~Yu, K.~Renz, and A.~Geiger.
\newblock Transfuser: Imitation with transformer-based sensor fusion for autonomous driving.
\newblock \emph{IEEE TPAMI}, 45\penalty0 (11):\penalty0 12878--12895, 2022.

\bibitem[Jaeger et~al.(2023)Jaeger, Chitta, and Geiger]{jaeger2023hidden}
B.~Jaeger, K.~Chitta, and A.~Geiger.
\newblock Hidden biases of end-to-end driving models.
\newblock In \emph{ICCV}, pages 8240--8249, 2023.

\bibitem[Huang et~al.(2020)Huang, Lv, Xing, and Wu]{huang2020multi}
Z.~Huang, C.~Lv, Y.~Xing, and J.~Wu.
\newblock Multi-modal sensor fusion-based deep neural network for end-to-end autonomous driving with scene understanding.
\newblock \emph{IEEE Sensors Journal}, 21\penalty0 (10):\penalty0 11781--11790, 2020.

\bibitem[Li et~al.(2018)Li, Motoyoshi, Sasaki, Ogata, and Sugano]{li2018rethinking}
Z.~Li, T.~Motoyoshi, K.~Sasaki, T.~Ogata, and S.~Sugano.
\newblock Rethinking self-driving: Multi-task knowledge for better generalization and accident explanation ability.
\newblock \emph{arXiv preprint arXiv:1809.11100}, 2018.

\bibitem[Xu et~al.(2017)Xu, Gao, Yu, and Darrell]{xu2017end}
H.~Xu, Y.~Gao, F.~Yu, and T.~Darrell.
\newblock End-to-end learning of driving models from large-scale video datasets.
\newblock In \emph{CVPR}, pages 2174--2182, 2017.

\bibitem[Chitta et~al.(2021)Chitta, Prakash, and Geiger]{chitta2021neat}
K.~Chitta, A.~Prakash, and A.~Geiger.
\newblock Neat: Neural attention fields for end-to-end autonomous driving.
\newblock In \emph{ICCV}, pages 15793--15803, 2021.

\bibitem[Zhang et~al.(2023)Zhang, Huang, and Ohn-Bar]{zhang2023coaching}
J.~Zhang, Z.~Huang, and E.~Ohn-Bar.
\newblock Coaching a teachable student.
\newblock In \emph{CVPR}, pages 7805--7815, 2023.

\bibitem[Shao et~al.(2023)Shao, Wang, Chen, Li, and Liu]{shao2023safety}
H.~Shao, L.~Wang, R.~Chen, H.~Li, and Y.~Liu.
\newblock Safety-enhanced autonomous driving using interpretable sensor fusion transformer.
\newblock In \emph{CoRL}, pages 726--737, 2023.

\bibitem[Jia et~al.(2023)Jia, Wu, Chen, Xie, He, Yan, and Li]{jia2023think}
X.~Jia, P.~Wu, L.~Chen, J.~Xie, C.~He, J.~Yan, and H.~Li.
\newblock Think twice before driving: Towards scalable decoders for end-to-end autonomous driving.
\newblock In \emph{CVPR}, pages 21983--21994, 2023.

\bibitem[Shao et~al.(2023)Shao, Wang, Chen, Waslander, Li, and Liu]{shao2023reasonnet}
H.~Shao, L.~Wang, R.~Chen, S.~L. Waslander, H.~Li, and Y.~Liu.
\newblock Reasonnet: End-to-end driving with temporal and global reasoning.
\newblock In \emph{CVPR}, pages 13723--13733, 2023.

\bibitem[Ishihara et~al.(2021)Ishihara, Kanervisto, Miura, and Hautamaki]{ishihara2021multi}
K.~Ishihara, A.~Kanervisto, J.~Miura, and V.~Hautamaki.
\newblock Multi-task learning with attention for end-to-end autonomous driving.
\newblock In \emph{CVPR}, pages 2902--2911, 2021.

\bibitem[Wu et~al.(2022)Wu, Jia, Chen, Yan, Li, and Qiao]{wu2022trajectory}
P.~Wu, X.~Jia, L.~Chen, J.~Yan, H.~Li, and Y.~Qiao.
\newblock Trajectory-guided control prediction for end-to-end autonomous driving: A simple yet strong baseline.
\newblock In \emph{NeurIPS}, volume~35, pages 6119--6132, 2022.

\bibitem[Xu et~al.(2023)Xu, Bazarjani, Chi, Choi, and Fu]{xu2023uncovering}
Y.~Xu, A.~Bazarjani, H.-g. Chi, C.~Choi, and Y.~Fu.
\newblock Uncovering the missing pattern: Unified framework towards trajectory imputation and prediction.
\newblock In \emph{CVPR}, pages 9632--9643, 2023.

\bibitem[Xu and Fu(2025)]{xu2025sportstraj}
Y.~Xu and Y.~Fu.
\newblock Sports-traj: A unified trajectory generation model for multi-agent movement in sports.
\newblock In \emph{ICLR}, 2025.

\bibitem[Li et~al.(2022)Li, Wang, Li, Xie, Sima, Lu, Qiao, and Dai]{li2022bevformer}
Z.~Li, W.~Wang, H.~Li, E.~Xie, C.~Sima, T.~Lu, Y.~Qiao, and J.~Dai.
\newblock Bevformer: Learning bird’s-eye-view representation from multi-camera images via spatiotemporal transformers.
\newblock In \emph{ECCV}, pages 1--18, 2022.

\bibitem[Loshchilov(2017)]{loshchilov2017decoupled}
I.~Loshchilov.
\newblock Decoupled weight decay regularization.
\newblock \emph{arXiv preprint arXiv:1711.05101}, 2017.

\bibitem[Loshchilov and Hutter(2016)]{loshchilov2016sgdr}
I.~Loshchilov and F.~Hutter.
\newblock Sgdr: Stochastic gradient descent with warm restarts.
\newblock \emph{arXiv preprint arXiv:1608.03983}, 2016.

\bibitem[Raffel et~al.(2020)Raffel, Shazeer, Roberts, Lee, Narang, Matena, Zhou, Li, and Liu]{raffel2020exploring}
C.~Raffel, N.~Shazeer, A.~Roberts, K.~Lee, S.~Narang, M.~Matena, Y.~Zhou, W.~Li, and P.~J. Liu.
\newblock Exploring the limits of transfer learning with a unified text-to-text transformer.
\newblock \emph{JMLR}, 21\penalty0 (140):\penalty0 1--67, 2020.

\bibitem[Song et~al.(2020)Song, Tan, Qin, Lu, and Liu]{song2020mpnet}
K.~Song, X.~Tan, T.~Qin, J.~Lu, and T.-Y. Liu.
\newblock Mpnet: Masked and permuted pre-training for language understanding.
\newblock In \emph{NeurIPS}, pages 16857--16867, 2020.

\bibitem[Kullback(1997)]{kullback1997information}
S.~Kullback.
\newblock \emph{Information theory and statistics}.
\newblock Courier Corporation, 1997.

\end{thebibliography}

\clearpage
\appendix
\section{Related Work}
\noindent
\textbf{Multi-Task Learning.}
Multi-task learning (MTL) jointly performs several related tasks using a shared representation through separate branches or heads. 
This approach leverages shared domain knowledge, enhancing feature robustness and generalization, making it well-suited for end-to-end autonomous driving.
In AD systems, auxiliary tasks such as semantic segmentation~\cite{chitta2022transfuser,jaeger2023hidden,huang2020multi,li2018rethinking,xu2017end}, depth estimation~\cite{li2018rethinking,xu2017end}, HD mapping, and BEV segmentation~\cite{chitta2021neat,zhang2023coaching,shao2023safety,jia2023think,shao2023reasonnet} are commonly adopted to extract meaningful perception representations for subsequent objects.
Beyond vision tasks, prior methods~\cite{ishihara2021multi,wu2022trajectory} have explored predicting auxiliary signals such as traffic light states and control commands, as well as unifying different trajectory modeling tasks~\cite{xu2023uncovering,xu2025sportstraj}, to enhance driving performance.
Inspired by the success of multi-task learning, 
we design novel auxiliary tasks to encourage the model to learn richer feature representations through high-quality reasoning annotations from VLMs, ultimately leading to more reliable planning performance.

\section{Implementation Details}
When integrating our proposed VLM-AD method into UniAD~\cite{hu2023planning}\footnote{\url{https://github.com/OpenDriveLab/UniAD}}, we follow the joint training protocol defined in UniAD. In the first stage, we initialize the model using BEVFormer~\cite{li2022bevformer} weights and train the perception and mapping tasks for 6 epochs. 
In the second stage, we freeze the image backbone and BEV encoder and perform end-to-end training using our proposed VLM-AD approach for 20 epochs. The model is trained using an initial learning rate of $2 \times 10^{-4}$, a learning rate multiplier of 0.1, and the AdamW optimizer~\cite{loshchilov2017decoupled} with a weight decay of 0.01.

When integrating our proposed VLM-AD method into VAD~\cite{jiang2023vad}\footnote{\url{https://github.com/hustvl/VAD}}, we adopt the same hyperparameters as in their original implementation. The model is trained using the AdamW optimizer~\cite{loshchilov2017decoupled} and a cosine annealing scheduler~\cite{loshchilov2016sgdr}, with a weight decay of 0.01 and an initial learning rate of $2 \times 10^{-4}$. 

When integrating our proposed VLM-AD method into SparseDrive~\cite{sun2024sparsedrive}\footnote{\url{https://github.com/swc-17/SparseDrive}}, we adopt the same hyperparameter settings as in their original implementation. Specifically, we use the SparseDrive-B base model with a ResNet-101 backbone and input image size of 512×1408. The model is trained using the AdamW optimizer~\cite{loshchilov2017decoupled} and a cosine annealing scheduler~\cite{loshchilov2016sgdr} in a two-stage training setup, with a learning rate of $3 \times 10^{-4}$ and a weight decay of $1 \times 10^{-3}$. 

For closed-loop evaluation, we integrate our method into the VAD-Base framework and follow the same experimental settings described in their paper~\cite{jiang2023vad}. Specifically, as in VAD-Base, the navigation information consists of a sparse goal location and a corresponding discrete navigational command, which are encoded using an MLP and passed to the planning head as part of the input features. Additionally, we retain the original traffic light classification branch used in VAD-Base, which comprises a ResNet-50 backbone followed by an MLP-based classification head. The input to this branch is the upper-middle region of the cropped front-view image, and the resulting feature map is flattened and provided to the planning head to help the model interpret traffic light states.

All experimental settings are kept consistent with the corresponding baseline models, as our proposed method is a plug-and-play module that can be seamlessly integrated into various end-to-end driving frameworks without architectural changes or hyperparameter tuning. This design ensures robustness, cross-model compatibility, and fair comparisons under consistent training conditions, while also highlighting the practicality of VLM-AD for real-world deployment.

To encode freeform annotations into text features, we use the pre-trained CLIP-ViT-B/32~\cite{radford2021learning} model with a dimension of 512. Additionally, we experiment with other text encoders such as T5-base~\cite{raffel2020exploring} and MPNet-base~\cite{song2020mpnet}, both encoding freeform annotations into text features with a dimension of 768, as described in Sec.~\ref{sec:ablation_study}.

\begin{table}[t!]
\centering
\caption{Results of different variants of VLM-KD. CLIP indicates the use of the contrastive learning loss defined in~\cite{radford2021learning} to align the text features of 
$Q_{1}$. MSE represents the use of MSE loss for feature alignment, KL represents KL divergence, and CosSim indicates cosine similarity for aligning features. Align refers to the alignment loss defined in our method.}
\begin{tabular}{l|c|cc}
\toprule
\multicolumn{1}{l|}{\multirow{2}{*}{Method}} & \multicolumn{1}{c|}{\multirow{2}{*}{Variant}} & \multicolumn{2}{c}{Planning Results } \\
& & Avg. L2 $\downarrow$ & Avg. Col $\downarrow$\\
\midrule

UniAD & - 
& 1.03 & 0.31\\

VLM-AD  & CLIP
& 0.94 & \underline{0.26} \\

VLM-AD  & MSE
& 0.99 & 0.30 \\

VLM-AD  & KL
& \underline{0.92} & \underline{0.26} \\

VLM-AD  & CosSim
& 0.96 & 0.28 \\

\rowcolor{RowColor} 
VLM-AD & Align 
& \textbf{0.89} & \textbf{0.24} \\

\bottomrule
\end{tabular}
\label{tab:contrastive}
\end{table}
\begin{table}[t!]
\centering
\caption{Results of different designs for VLM-KD. MLP indicates that an MLP layer is used to replace the MHCA block for $Q_{2}$. T5 and MPNet indicate the use of different language models to convert reasoning annotations from $Q_{1}$ into reasoning features that serve as supervision labels.}
\begin{tabular}{l|c|cc}
\toprule
\multicolumn{1}{l|}{\multirow{2}{*}{Method}} & \multicolumn{1}{c|}{\multirow{2}{*}{Variant}} & \multicolumn{2}{c}{Planning Results } \\
& & Avg. L2 $\downarrow$ & Avg. Col $\downarrow$\\
\midrule

UniAD & - 
& 1.03 & 0.31\\

VLM-AD & MLP
& 0.94 & 0.29  \\

\rowcolor{RowColor} 
VLM-AD & MHCA
& \textbf{0.91} & \textbf{0.29} \\

\midrule

VLM-AD & T5
& 0.94 & 0.29  \\

VLM-AD & MPNet
& 0.91 & 0.26 \\

\rowcolor{RowColor} 
VLM-AD & CLIP
& \textbf{0.89} & \textbf{0.24} \\



\bottomrule
\end{tabular}
\label{tab:design}
\end{table}
\begin{table}[t!]
\centering
\caption{Results of using different hyperparameters of $\lambda_{1}$ and $\lambda_{2}$ to control the weights of $\mathcal{L}_{align}$ and $\mathcal{L}_{action}$.}
\scalebox{0.95}{
\begin{tabular}{l|c|cc}
\toprule
\multicolumn{1}{l|}{\multirow{2}{*}{Method}} & \multicolumn{1}{c|}{\multirow{2}{*}{Variant}} & \multicolumn{2}{c}{Planning Results} \\
& & Avg. L2 $\downarrow$ & Avg. Col $\downarrow$\\
\midrule

UniAD & - 
& 1.03 & 0.31\\

VLM-AD & $\lambda_{1}=1$, $\lambda_{2}=1$
& \underline{0.90} & \underline{0.26} \\

VLM-AD & $\lambda_{1}=0.1$, $\lambda_{2}=1$
& 0.92 & 0.29\\

\rowcolor{RowColor} 
VLM-AD & $\lambda_{1}=1$, $\lambda_{2}=0.1$
& \textbf{0.88} & \textbf{0.19} \\

\bottomrule
\end{tabular}
}
\label{tab:hyper}
\end{table}

\section{Ablation Study}
\label{sec:appendix_ablation_study}
\noindent
\textbf{Contrastive Learning with Text Annotations.}
In addition to our proposed auxiliary task, we employ a contrastive learning strategy similar to VLP. We follow VLP~\cite{pan2024vlp} to project our generated annotations from $Q_{1}$ using CLIP~\cite{radford2021learning} to obtain three distinct text features. Each text feature is then contrasted with the ego feature individually before we compute the contrastive loss.

The results, presented in \cref{tab:contrastive}, show that contrastive learning improves performance compared to the baseline, though it is still worse than our custom auxiliary tasks. This may be because each frame contains only one ego vehicle, resulting in a single positive sample for contrastive learning that can lead to unstable alignment.

\noindent
\textbf{Feature Alignment Loss.}
We also study alternative options of feature alignment, including minimizing contrastive learning loss in CLIP~\cite{radford2021learning}, MSE loss, KL divergence loss~\cite{kullback1997information}, or maximizing negative cosine similarity to align the three features from $Q{1}$. The results, shown in \cref{tab:contrastive}, indicate that MSE loss performs slightly better than UniAD, by minimizing the Euclidean distance between features, driving outputs toward their mean, which causes information loss during training. Both CLIP loss, KL divergence, and cosine similarity outperform UniAD but are inferior to our alignment loss.
This underscores the importance of normalizing with different temperatures to balance the smoothness and sharpness of teacher-student features.

\noindent
\textbf{Model Design.}
We investigate alternative design options in our method. First, we use an MLP layer instead of the MHCA block in our structured action classification head. Second, we study different language models, such as T5~\cite{raffel2020exploring}, MPNet~\cite{song2020mpnet} in addition to CLIP for encoding text annotations from $Q_{1}$. From \cref{tab:design}, we observe that MLP achieves slightly worse L2 performance and the same collision rate. Additionally, we observe that both T5 and MPNet outperform the UniAD baseline, but are slightly worse than CLIP.

\noindent
\textbf{Hyperparameter Study.}
Balancing the losses of different tasks is a critical challenge in multi-task learning. We study the hyperparameters $\lambda_{1}$ and $\lambda_{2}$ in the context of UniAD. 
The results, shown in \cref{tab:hyper}, indicate that all three variants outperform UniAD. 
Among these variants, the performance is the worst when
$\lambda_{1}=0.1$ and $\lambda_{2}=1$, as the annotations for $Q_{1}$ contain more valuable information compared to the annotations for $Q_{2}$.

\begin{figure}[t!]
  \centering
  \begin{subfigure}[t]{0.48\linewidth}
    \centering
    \includegraphics[width=\linewidth]{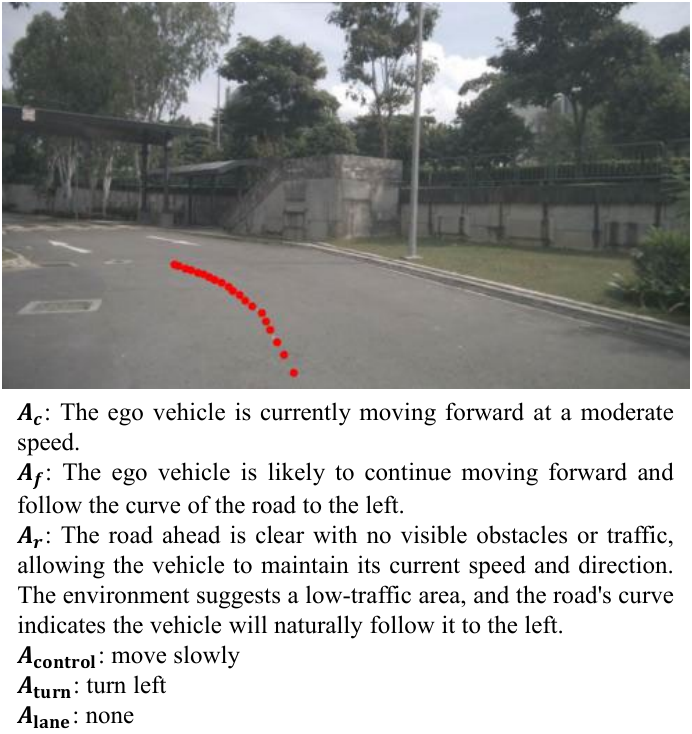}
    \caption{Annotation using a single front-view image.}
    \label{fig:proj}
  \end{subfigure}
  \hfill
  \begin{subfigure}[t]{0.48\linewidth}
    \centering
    \includegraphics[width=\linewidth]{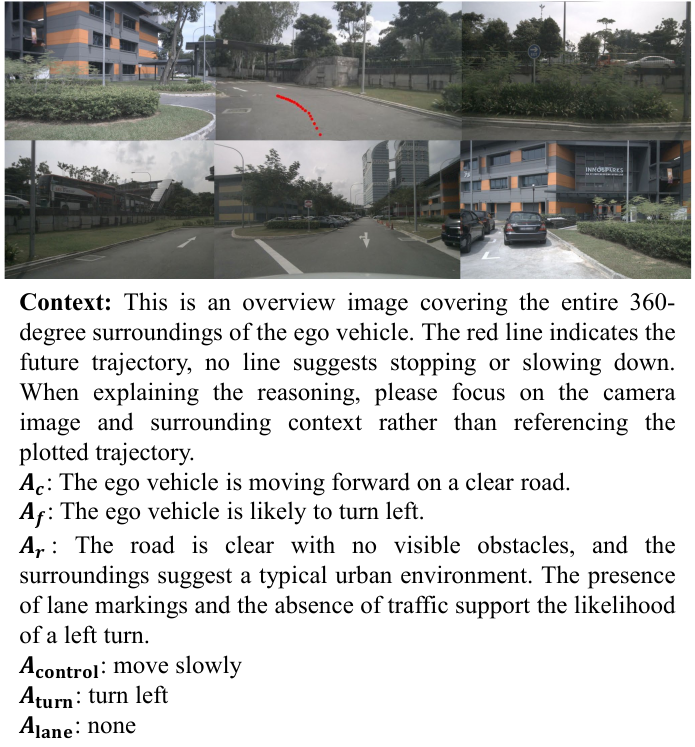}
    \caption{Annotation using all 6 images.}
    \label{fig:all}
  \end{subfigure}
  \caption{Comparison of annotation strategies: (a) only the front-view image is used; (b) all 6 camera views are included, but the trajectory is still projected onto the front view.}
  \label{fig:anno_compare}
\end{figure}

\begin{figure*}[t!]
  \centering
   \includegraphics[width=0.98\linewidth]{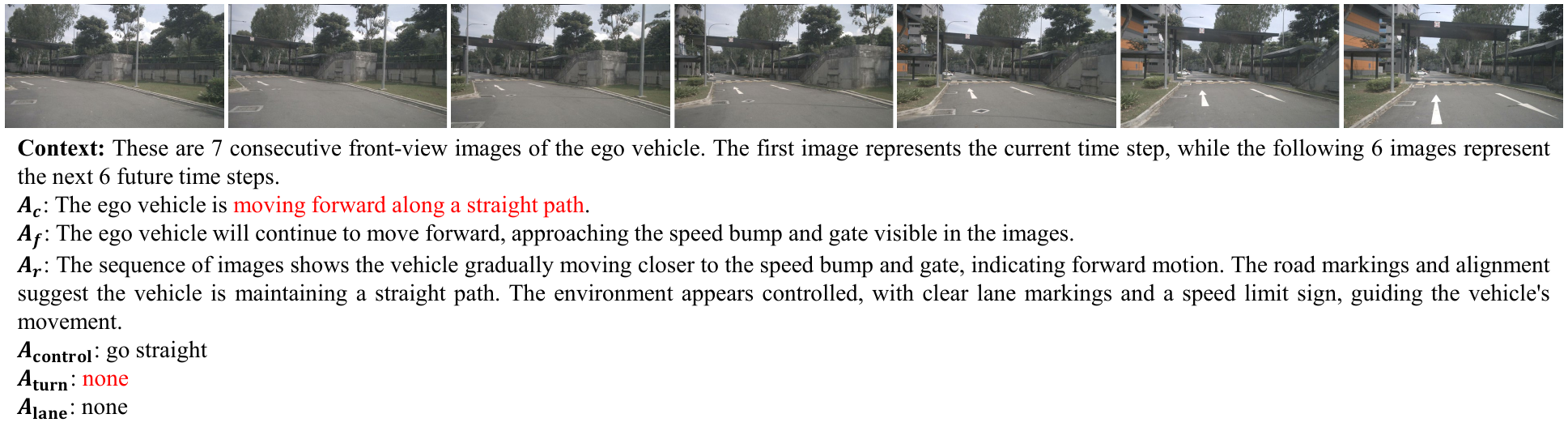}
   \caption{An annotation example using 7 consecutive front-view images as the visual input, which results in incorrect annotations.}
   \label{fig:consecutive}
\end{figure*}

\section{VLM Annotation}
In this section, we provide a comprehensive analysis of our VLM annotations, including visual input format comparisons, annotation statistics, as well as the annotation quality, including representative successful and failure examples.

\subsection{Visual Input}
While we use a front-view image (as shown in \cref{fig:proj}) as the visual input to the VLM, we also experiment with other alternatives described in \cref{sec:method}, including six images covering the full 360-degree surroundings of the ego vehicle (\cref{fig:all}) and a sequence of consecutive front-view images (\cref{fig:consecutive}).
Compared to the first alternative that uses the full 360-degree surrounding as input, our approach yields similar annotations from the VLM while significantly reducing the computational cost by processing a much smaller input image.
The second alternative, which utilizes consecutive front-view images, often results in incorrect annotations, such as misidentifying the current action status and failing to detect a left-turn action. 
This is due to the challenge VLMs face in understanding temporal dynamics, especially from ego-centric visual signals. Furthermore, using consecutive images increases the annotation time by approximately 80\% compared to our approach.

\begin{table}[t!]
\centering
\caption{Statistics of freeform text annotation $\mathcal{A}_{1}$ from $Q_{1}$.}
\scalebox{0.8}{
\begin{tabular}{l|ccc}
\toprule
\multicolumn{1}{l|}{\multirow{2}{*}{Word Length}} & \multicolumn{3}{c}{Annotation $\mathcal{A}_{1}$} \\
\cmidrule(lr){2-4}
& $\mathcal{A}_{c}$ & $\mathcal{A}_{f}$ & $\mathcal{A}_{r}$ \\
\midrule
Max & 29 & 41 & 93\\ 
Min & 6 & 7 & 13\\ 
Mean & 11.30 & 15.34 & 35.85\\ 
\bottomrule
\end{tabular}
}
\label{tab:text_sta}
\end{table}

\begin{figure}[t!]
  \centering
  \begin{subfigure}[t]{0.48\linewidth}
    \centering
    \includegraphics[width=\linewidth]{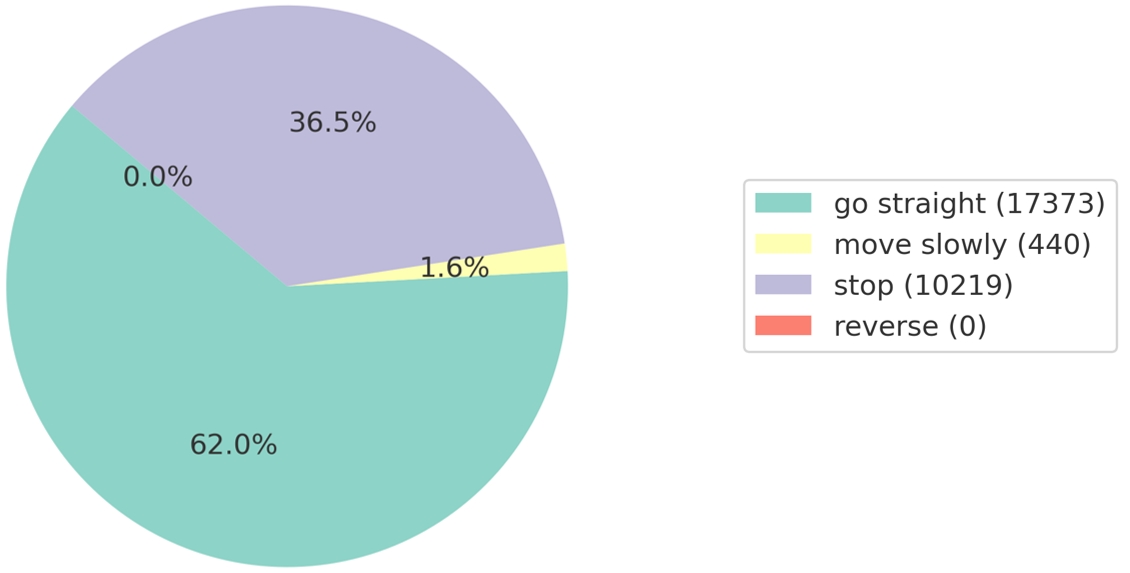}
    \caption{Control action}
    \label{fig:control}
  \end{subfigure}
  \hfill
  \begin{subfigure}[t]{0.48\linewidth}
    \centering
    \includegraphics[width=\linewidth]{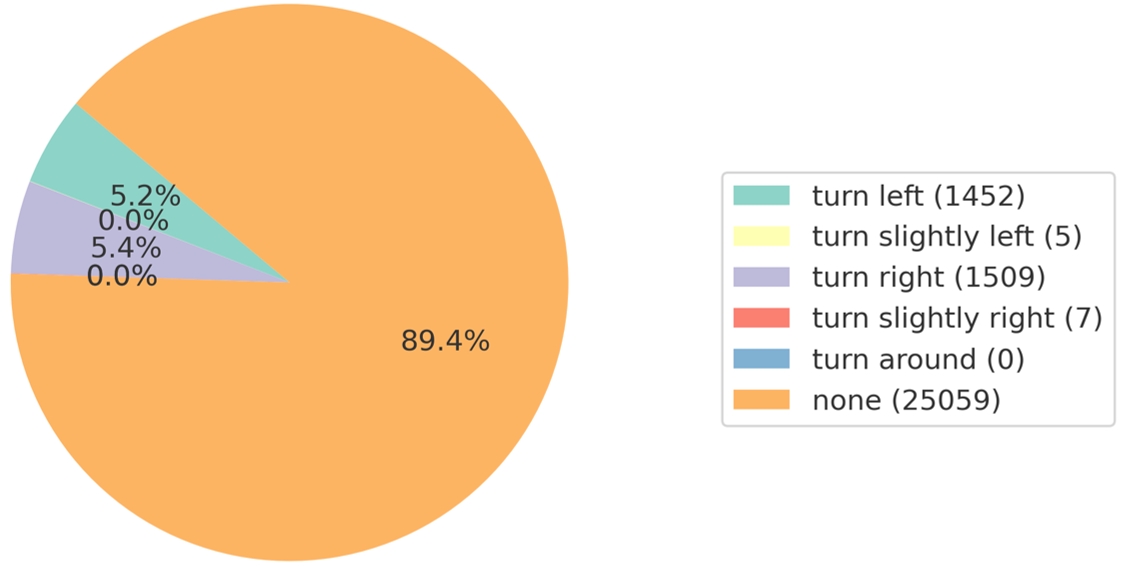}
    \caption{Turn action}
    \label{fig:turn}
  \end{subfigure}
  
  \vspace{2mm}
  
  \begin{subfigure}[t]{0.48\linewidth}
    \centering
    \includegraphics[width=\linewidth]{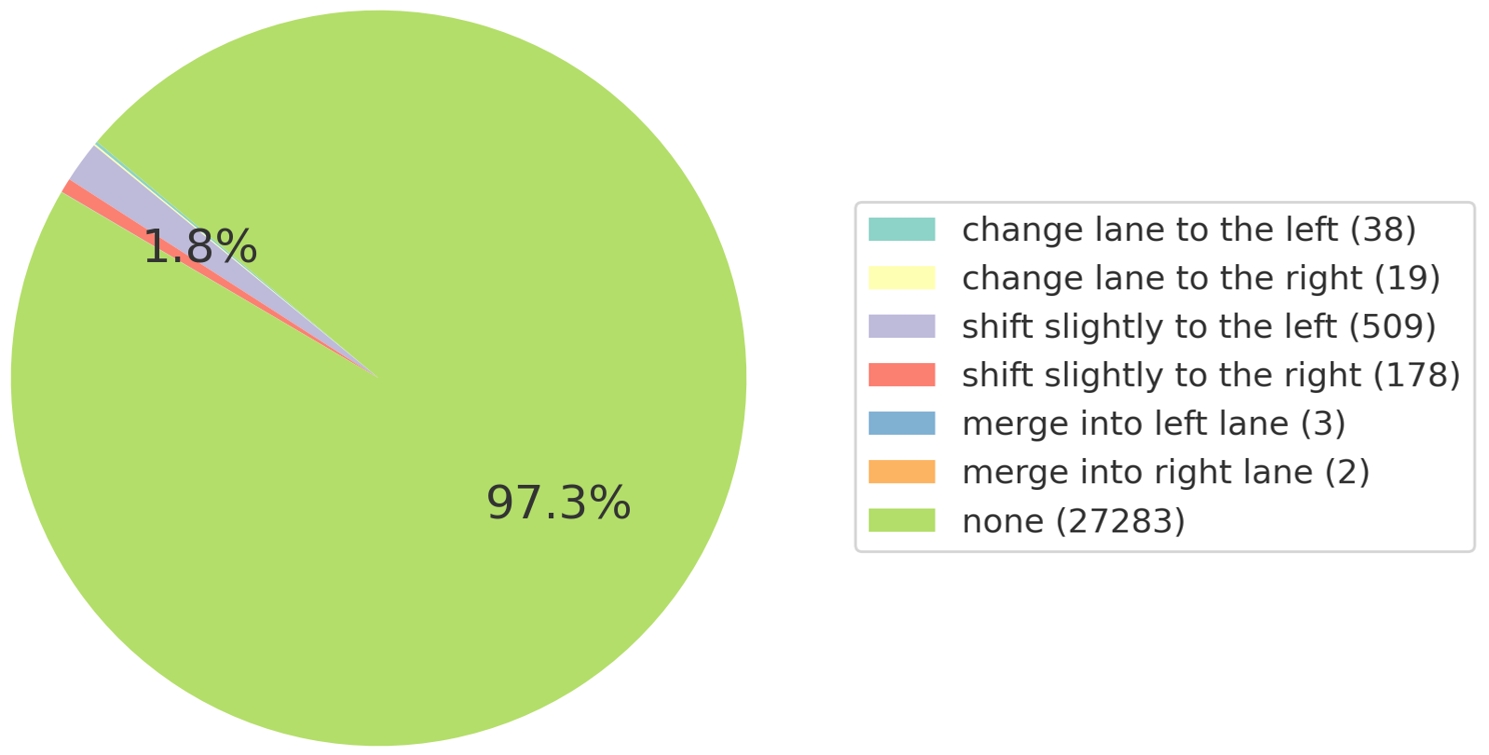}
    \caption{Lane action}
    \label{fig:lane}
  \end{subfigure}
  \caption{Distributions of control, turn, and lane actions.}
  \label{fig:action_dist}
\end{figure}

\subsection{Annotation Statistics}
We annotate the training set of the nuScenes dataset, which consists of 700 scenes and 28,130 frames. Following the methodology described in \cref{sec:method}, we set $T=6$ to project the ego vehicle's future trajectory onto the front-view image. Consistent with UniAD~\cite{hu2023planning}, we exclude samples lacking sufficient input data, resulting in a total of 28,032 annotated samples.

For freeform reasoning annotations using $Q_{1}$, we calculate the word length of responses for each sub-question ($Q_{1-1}$, $Q_{1-2}$, and $Q_{1-3}$). The statistics are presented in \cref{tab:text_sta}, in which the average response length of $\mathcal{A}_{r}$ is the longest, as this sub-question focuses on detailed reasoning information.

For structured action annotations using $Q_{2}$, we analyze the distribution of three types of actions. The results are shown in \cref{fig:control}, \cref{fig:turn} and \cref{fig:lane}.
Approximately 62\% of frames are labeled as ``go straight'', 89.4\% as ``no turn action'', and 97.3\% as ``no lane action''. Notably, no frames are labeled with ``reverse'' or ``turn around'', and only a very small number of frames are labeled as ``merge into left lane'' or ``merge into right lane''. These statistics suggest that the nuScenes dataset has limited diversity in driving actions.

An interesting observation is that the VLM occasionally outputs actions not included in our predefined action list. 
For example, it generates actions such as ``turn slightly left'', ``turn slightly right'', ``shift slightly to the left'', and ``shift slightly to the right''. 
In our work, we merge these outputs into our predefined one-hot categories: ``turn slightly left'' is merged with ``turn left'', ``turn slightly right'' with ``turn right'', ``shift slightly to the left'' with ``change lane to the left'', and ``shift slightly to the right'' with ``change lane to the right''. This highlights the advantage of using structured annotations, as they help mitigate hallucinations by constraining VLM outputs to predefined categories.

\begin{table}[t!]
\centering
\caption{Summary of questionnaire results from 5 participants evaluating both freeform annotations and structured action labels. Std represents the standard deviation among the participants.}
\begin{tabular}{l|ccc|ccc}
\toprule
\multicolumn{1}{l|}{\multirow{2}{*}{Participant}} & \multicolumn{3}{c}{Average Score (1-5)} & \multicolumn{3}{c}{Accuracy (\%)}\\
\cmidrule(lr){2-7}
& $\mathcal{A}_{c}$ & $\mathcal{A}_{f}$ & $\mathcal{A}_{r}$ 
& $\mathcal{A}_{\text{control}}$ & $\mathcal{A}_{\text{turn}}$ & $\mathcal{A}_{\text{lane}}$\\
\midrule
1 & 4.58 & 4.66 & 4.26 & 0.88 & 0.96 & 0.98 \\
2 & 4.34 & 4.26 & 4.34 & 0.86 & 0.92 & 0.94 \\	
3 & 4.86 & 4.66 & 4.54 & 0.98 & 0.84 & 0.96 \\	
4 & 4.12 & 4.34 & 4.40 & 0.80 & 0.84 & 0.94 \\	
5 & 4.50 & 4.62 & 4.56 & 0.98 & 0.96 & 0.98 \\
\midrule
\textbf{Average} & 4.48 & 4.51 & 4.42 & 0.90 & 0.90 & 0.96 \\
\textbf{Std} & 0.28 & 0.19 & 0.13 & 0.08 & 0.06	 & 0.02 \\
\bottomrule
\end{tabular}
\label{tab:quality}
\end{table}





\subsection{Annotation Quality}
To validate the annotation quality from the VLM, we make a questionnaire with a random sample of 50 cases for evaluation. For each case, participants are provided with the front-view image of the ego vehicle, projected with its future trajectory, along with the corresponding VLM annotations of $Q_{1}$ and $Q_{2}$.

Participants are then asked to score each response. For freeform reasoning annotations, we set a scoring criterion on a 1 to 5 scale as follows:
\begin{itemize}[leftmargin=*, labelsep=0.5em, itemindent=0pt]
    \item \textit{5 Points}: Highly Consistent.
        \begin{itemize}[label=$\circ$]
            \item The text description perfectly matches the image.
            \item Key elements of the image (e.g., vehicle state, action, reasoning) are accurately described. 
            \item The text is clear, concise, and complete, with no unnecessary details or contradictions.
        \end{itemize}
    \item \textit{4 Points}: Mostly Consistent.
        \begin{itemize}[label=$\circ$]
            \item The text description mostly aligns with the image, with minor inaccuracies or omissions.
            \item Key elements are described but may lack some secondary details.
            \item Alternatively, the text may contain minor redundancies or slightly unrelated details that don’t impact the overall match.
        \end{itemize}
    \item \textit{3 Points}: Partially Consistent.
        \begin{itemize}[label=$\circ$]
            \item The text description partially matches the image but has notable inaccuracies or missing details.
            \item Important aspects of the image (e.g., vehicle speed, road conditions) may be under- or misrepresented.
            \item There could be some conflicting or vague statements.
        \end{itemize}
    \item \textit{2 Points}: Mostly Inconsistent.
        \begin{itemize}[label=$\circ$]
            \item The text description is largely inconsistent with the image but contains a small amount of relevant information.
            \item The description fails to capture critical details of the image or includes noticeable inaccuracies.
            \item Logical errors or contradictions in the text are present.
        \end{itemize}
    \item \textit{1 Point}: Completely Inconsistent.
        \begin{itemize}[label=$\circ$]
            \item The text description does not match the image at all.
            \item The text is entirely irrelevant or contradicts the image in significant ways.
            \item Misleading information that severely detracts from interpretability.
        \end{itemize}
\end{itemize}

For structured action annotations, we ask the participants to score True or False for each action.

We evaluated the results from 5 participants, as summarized in \cref{tab:quality}. The scores validate the overall annotation quality. Specifically, the annotation $\mathcal{A}_{f}$, which predicts future actions, received the highest score, while the annotation $\mathcal{A}_{r}$, which describes reasoning, received the lowest. Additionally, for action annotations, the accuracy for all three action types is 90\% or higher, with the lane action achieving the highest accuracy at 96\%.

\subsection{Succesful Annotation Examples}
\begin{figure}[t!]
  \centering
  \begin{subfigure}[t]{0.48\linewidth}
    \centering
    \includegraphics[width=\linewidth]{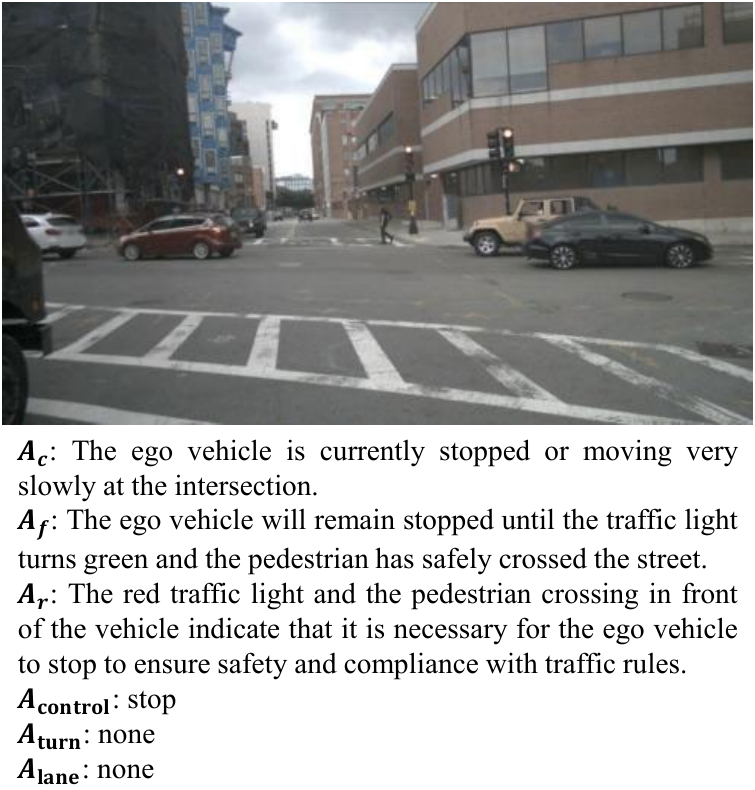}
    \caption{A successful annotation example where the red traffic light is correctly identified, and the annotations are complete and informative.}
    \label{fig:case1}
  \end{subfigure}
  \hfill
  \begin{subfigure}[t]{0.48\linewidth}
    \centering
    \includegraphics[width=\linewidth]{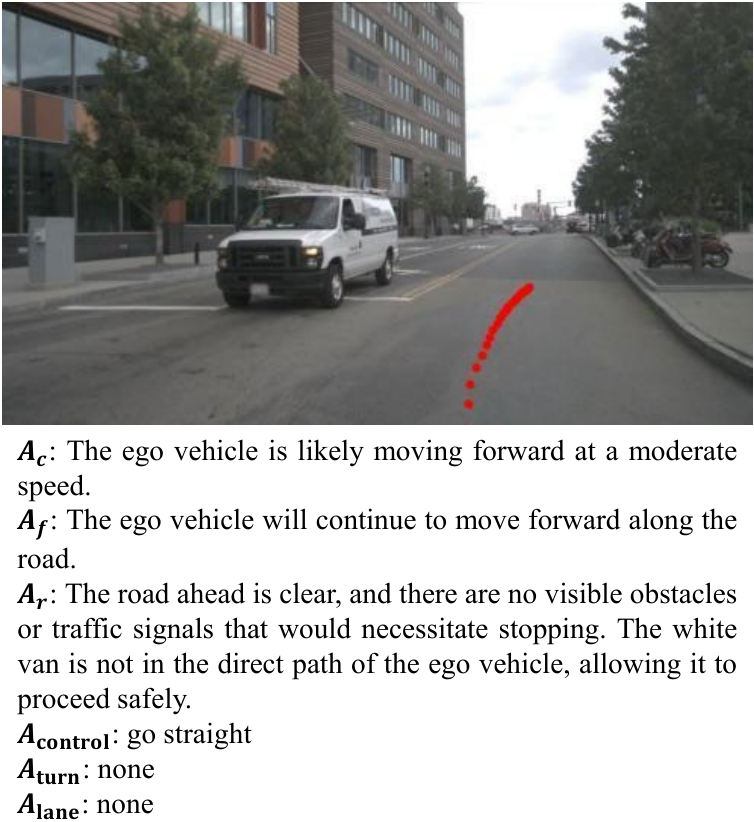}
    \caption{A successful annotation example where the white van is correctly identified as irrelevant to the ego vehicle's driving.}
    \label{fig:case2}
  \end{subfigure}

    \vspace{10mm}
  
  \begin{subfigure}[t]{0.48\linewidth}
    \centering
    \includegraphics[width=\linewidth]{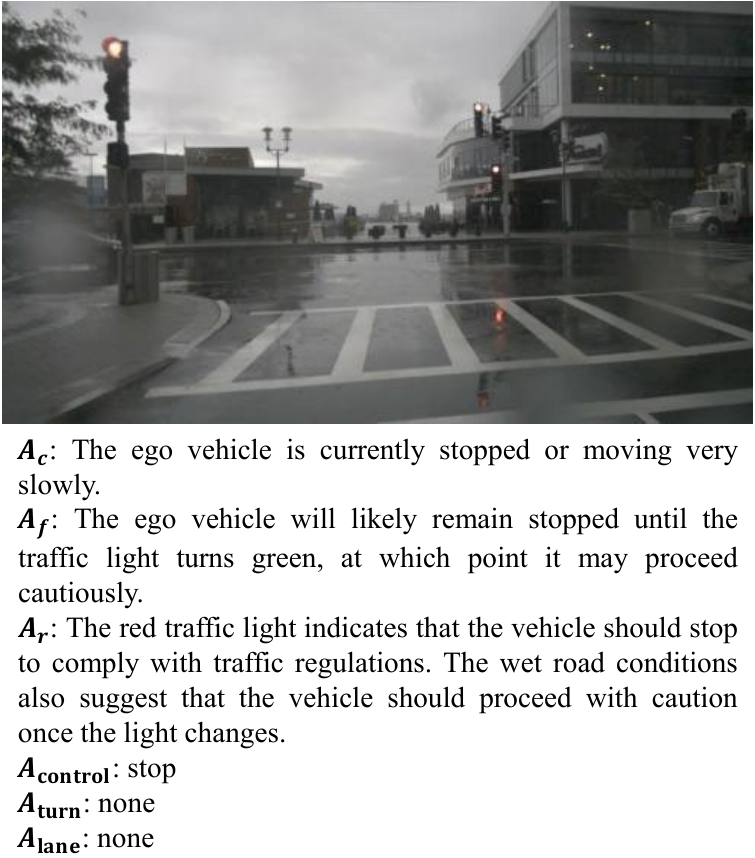}
    \caption{A successful annotation example where the rainy condition and red traffic light are correctly identified, resulting in high-quality annotations.}
    \label{fig:case3}
  \end{subfigure}
  \caption{Qualitative examples of successful VLM-based annotations.}
  \label{fig:success_cases}
\end{figure}

We present three examples to demonstrate the quality of VLM annotations, as shown in \cref{fig:case1}, \cref{fig:case2}, and \cref{fig:case3}.

In \cref{fig:case1}, the VLM accurately identifies the red traffic light and suggests a stop action at the intersection. It also predicts a reasonable future action and clearly explains the rationale behind its decisions.

In \cref{fig:case2}, a white van is observed ahead of the ego vehicle but in the opposite lane. The VLM correctly assesses that the van will not affect the ego vehicle's movement and outputs appropriate driving actions.

In \cref{fig:case3}, the ego vehicle is stopped at an intersection on a rainy day. Despite low visibility, the VLM successfully identifies the red traffic light and predicts the future movement based on the traffic light’s status.

\subsection{Imperfect Annotation Examples}
\begin{figure}[t!]
  \centering
  \begin{subfigure}[t]{0.48\linewidth}
    \centering
    \includegraphics[width=\linewidth]{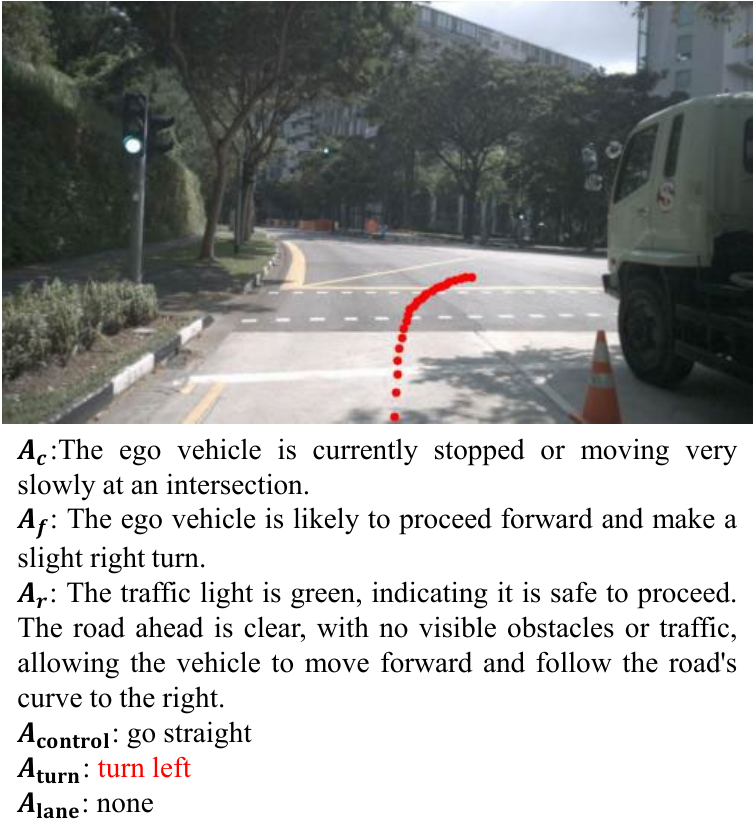}
    \caption{An imperfect annotation example where reasoning annotations are accurate but the turn action is incorrectly annotated as ``turn left''.}
    \label{fig:f_case1}
  \end{subfigure}
  \hfill
  \begin{subfigure}[t]{0.48\linewidth}
    \centering
    \includegraphics[width=\linewidth]{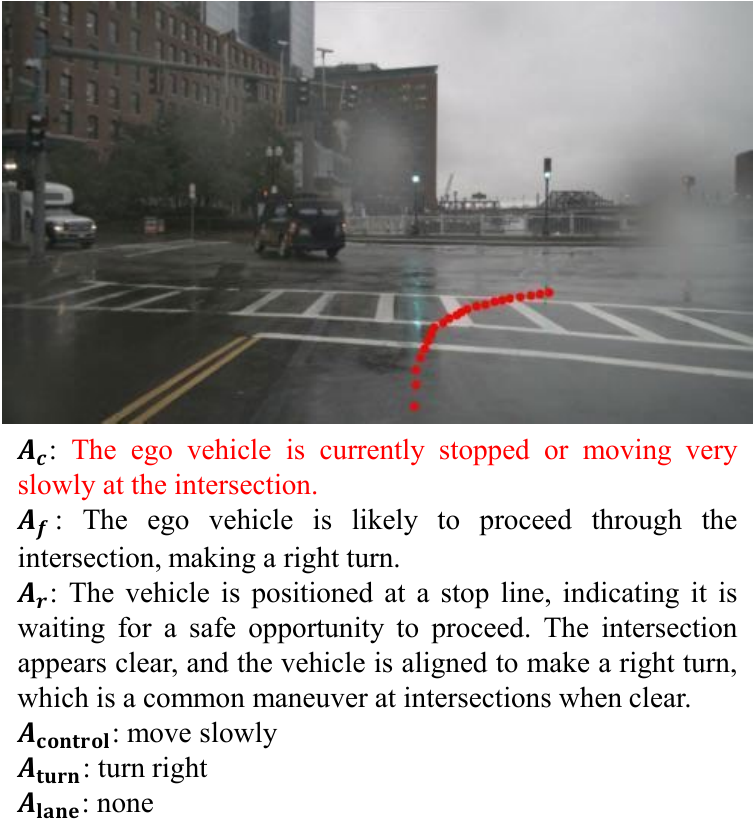}
    \caption{An imperfect annotation example where an overly cautious annotation is generated, despite the ego vehicle continuing without stopping or slowing down.}
    \label{fig:f_case2}
  \end{subfigure}

    \vspace{10mm}
  
  \begin{subfigure}[t]{0.48\linewidth}
    \centering
    \includegraphics[width=\linewidth]{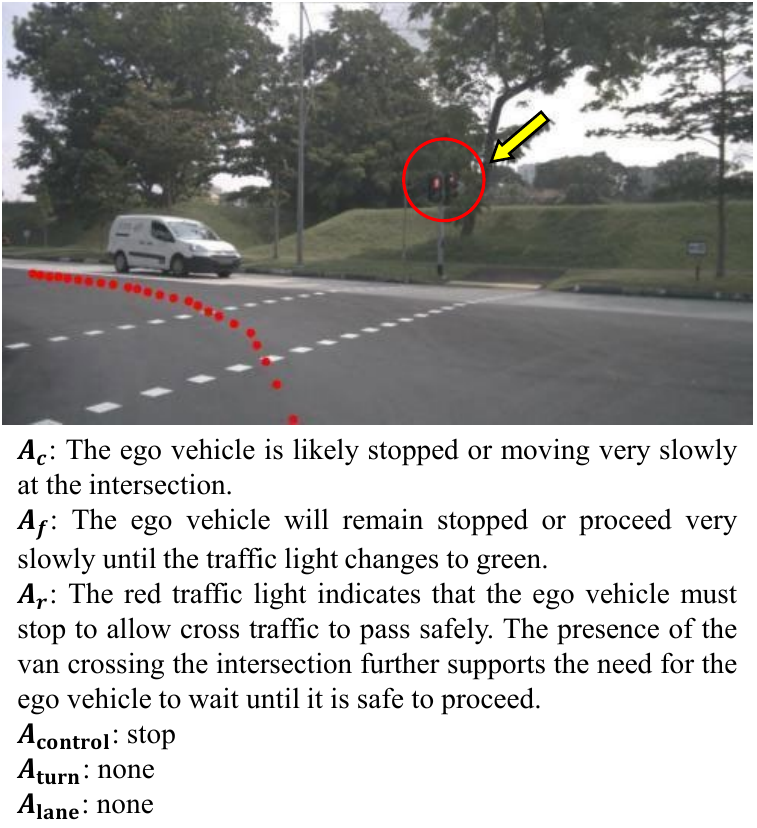}
    \caption{An imperfect annotation example where the pedestrian traffic light is mistakenly considered, leading to an incorrect stop annotation.}
    \label{fig:f_case3}
  \end{subfigure}
  \caption{Qualitative examples of imperfect VLM-based annotations.}
  \label{fig:failure_cases}
\end{figure}

We also present three annotation failure cases, illustrated in \cref{fig:f_case1}, \cref{fig:f_case2}, and \cref{fig:f_case3}.

In \cref{fig:f_case1}, the VLM accurately recognizes that the traffic light is green and predicts a future right-turn action from its reasoning annotation. However, it incorrectly outputs a left-turn action from the action annotation. Since we query $Q_{1}$ and $Q_{2}$ separately, the response to $Q_{1}$ does not influence $Q_{2}$. 
One potential solution is to introduce additional prompts to establish a progressive questioning process toward more accurate action annotations.

In \cref{fig:f_case2}, the VLM outputs ``stop'' or ``move slowly'' as the ego vehicle's current status. While these outputs are plausible, they are inconsistent with the ground truth, as the projected future trajectory indicates that the ego vehicle is currently turning right at the intersection. On the other hand, the action annotation successfully predicts the correct future action.

In \cref{fig:f_case3}, the VLM mistakenly identifies the red pedestrian light as a traffic light and provides incorrect responses.

Overall, despite making occasional mistakes, the VLM demonstrates the ability to generate meaningful annotations that uncover the underlying rationale behind driving decisions, bringing benefits to our proposed method as validated by our experiments.
By querying two independent annotation questions, our method is robust to VLM mistakes that often appear in only one of the two responses, as seen in~\cref{fig:f_case1} and~\cref{fig:f_case2}.
We defer to obtaining more accurate VLM responses as future work to further advance the performance of E2E planning models.

\section{Qualitative Examples}
\begin{figure*}[ht]
  \centering
   \includegraphics[width=0.98\linewidth]{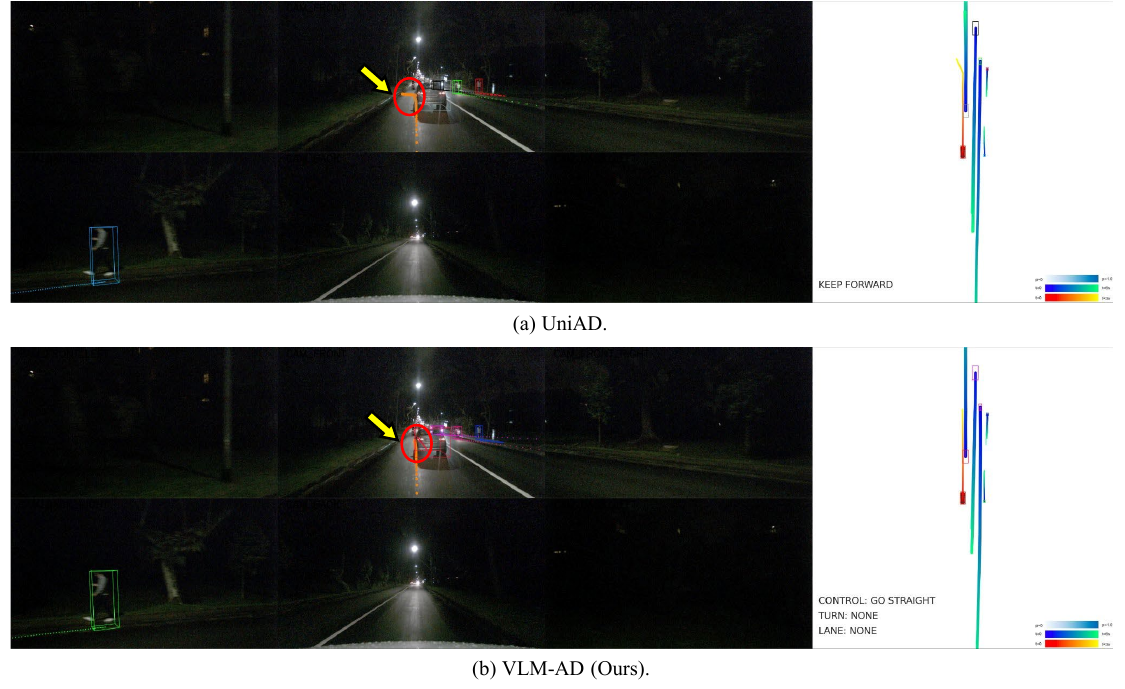}
   \caption{Our method generates a smooth trajectory for an evening driving scenario, in contrast to the baseline method that predicts a winding trajectory.}
   \label{fig:supp_vis_1}
\end{figure*}

\begin{figure*}[ht]
  \centering
   \includegraphics[width=0.98\linewidth]{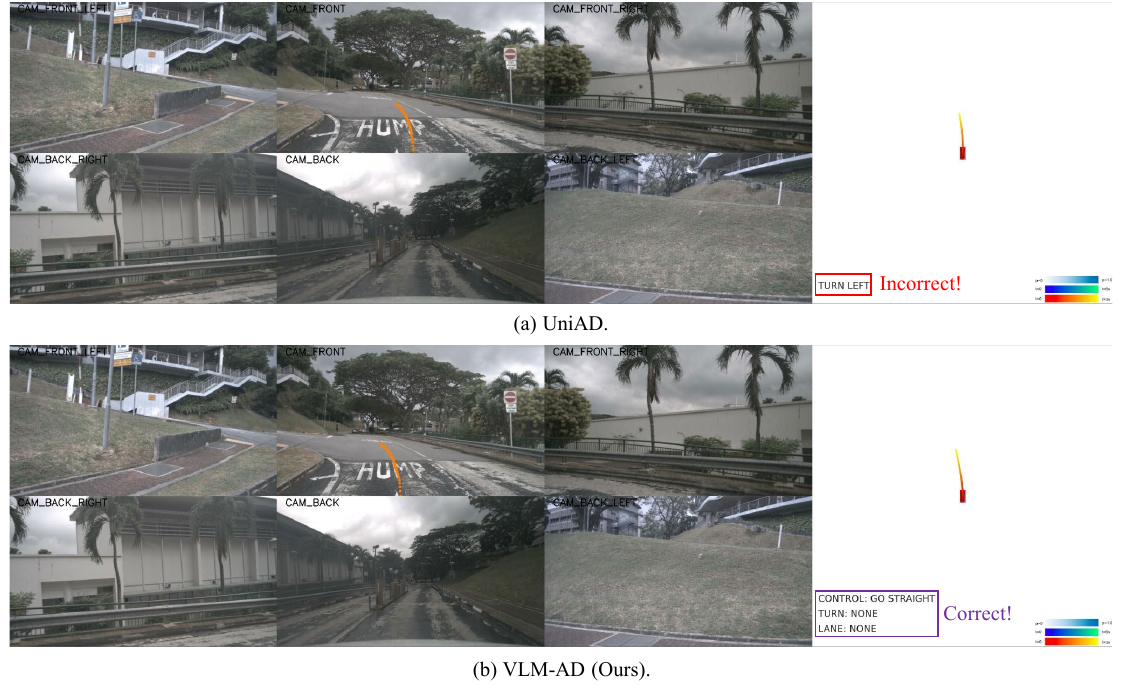}
   \caption{Our method accurately predicts the correct actions when driving on a curved road, while the command generated by UniAD is incorrect.}
   \label{fig:supp_vis_2}
\end{figure*}

\begin{figure*}[ht]
  \centering
   \includegraphics[width=0.98\linewidth]{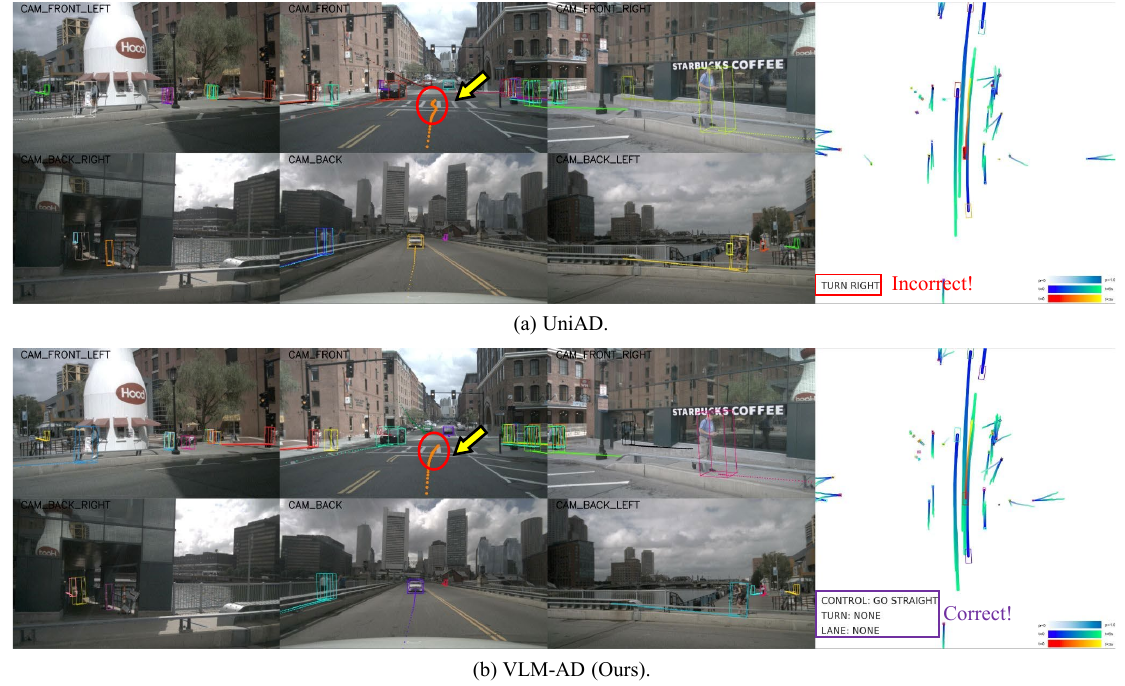}
   \caption{Our method predicts accurate actions and a smooth trajectory in an urban driving scenario, in contrast to the baseline method that predicts a winding trajectory based on an incorrect command.}
   \label{fig:supp_vis_3}
\end{figure*}

\begin{figure*}[ht]
  \centering
   \includegraphics[width=0.98\linewidth]{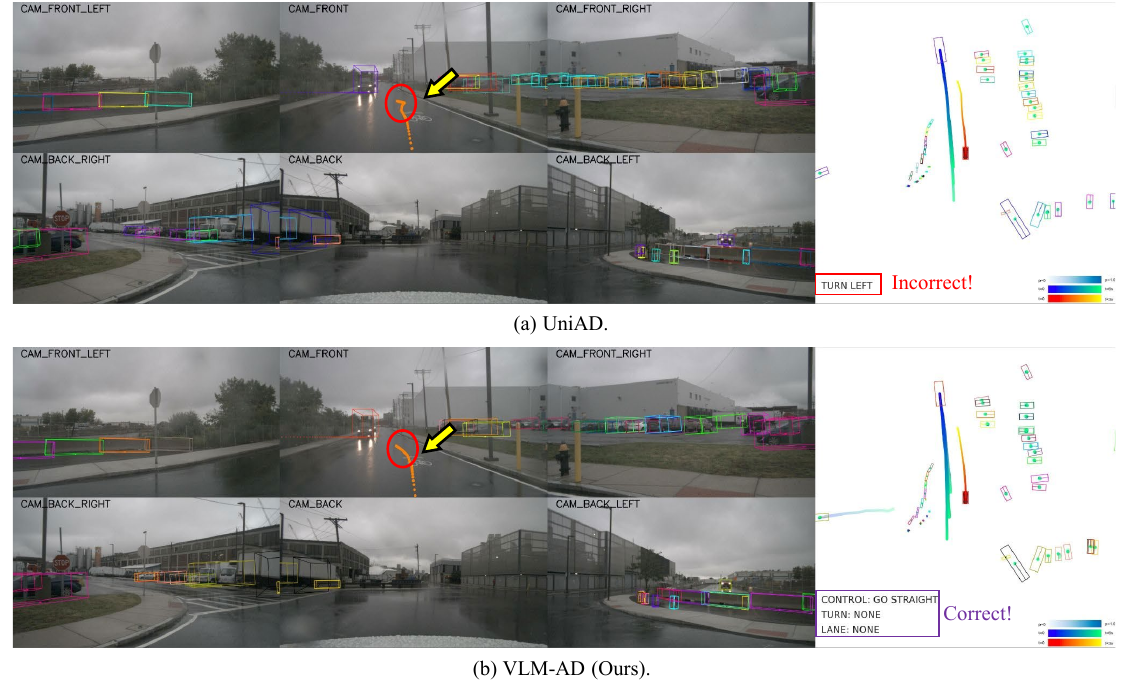}
   \caption{Our method predicts accurate actions and a smooth trajectory in rainy conditions, in contrast to the baseline method that predicts a winding trajectory based on an incorrect command.}
   \label{fig:supp_vis_4}
\end{figure*}

\begin{figure*}[ht]
  \centering
   \includegraphics[width=0.98\linewidth]{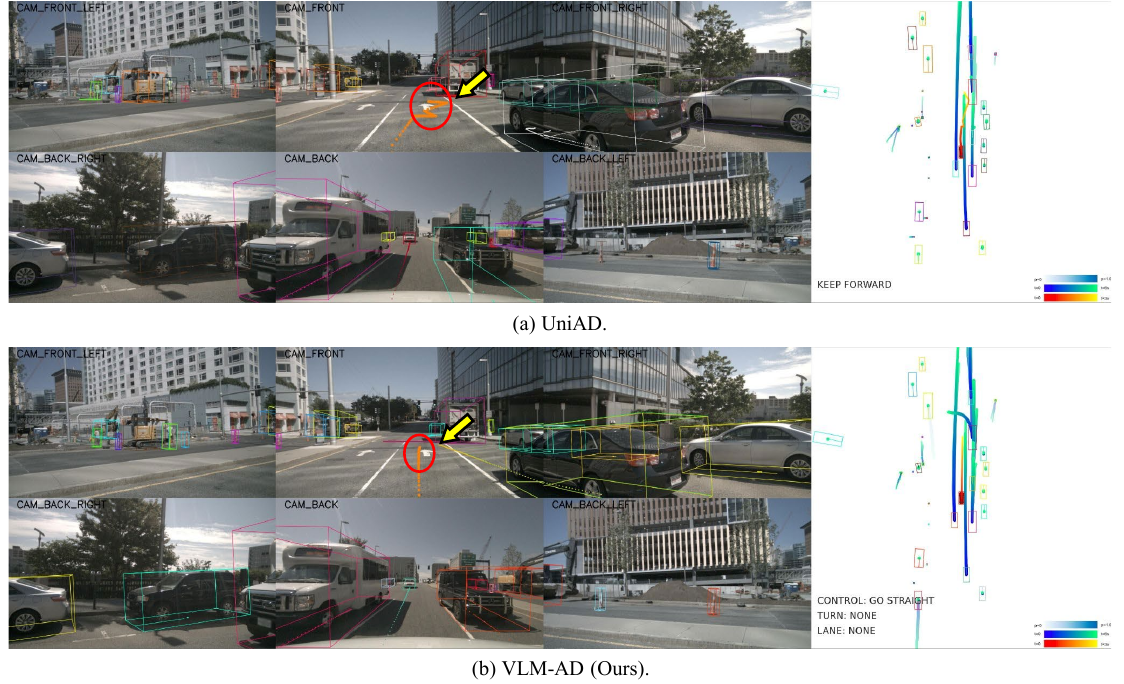}
   \caption{Our method predicts an accurate trajectory that stays within the lane boundaries, in contrast to the baseline method that predicts a zigzagging trajectory.}
   \label{fig:supp_vis_5}
\end{figure*}

\begin{figure*}[ht]
  \centering
   \includegraphics[width=0.98\linewidth]{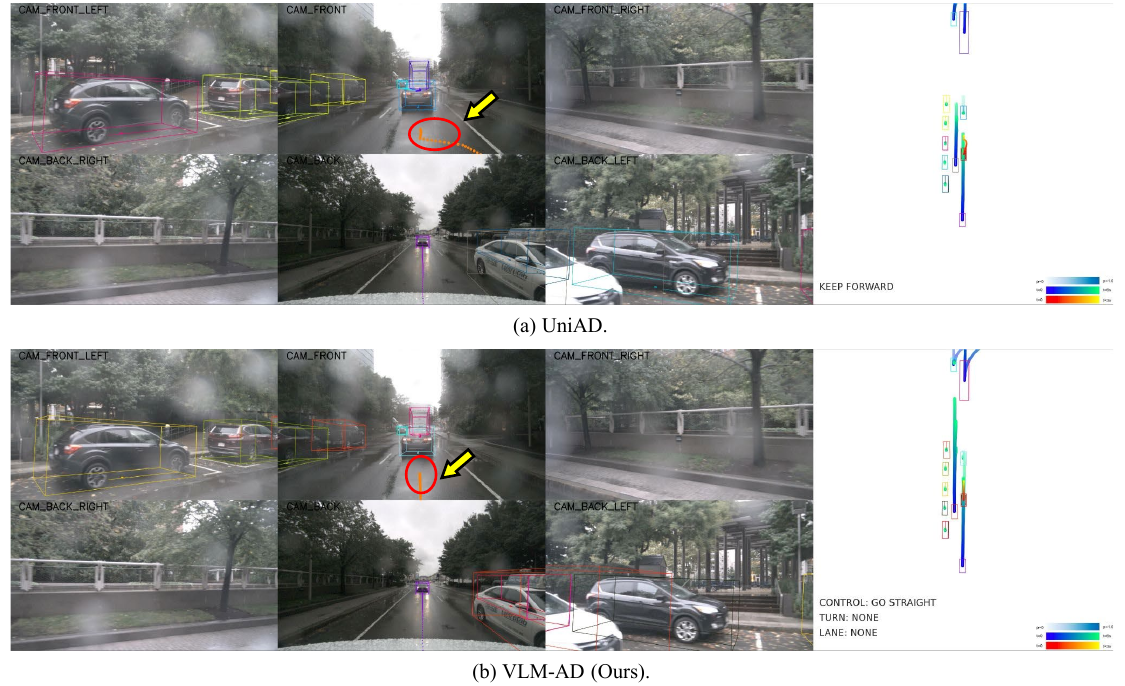}
   \caption{Our method predicts a smooth trajectory that stays within the lane boundaries, in contrast to the baseline method that predicts a swerving trajectory.}
   \label{fig:supp_vis_6}
\end{figure*}

\begin{figure*}[ht]
  \centering
   \includegraphics[width=0.98\linewidth]{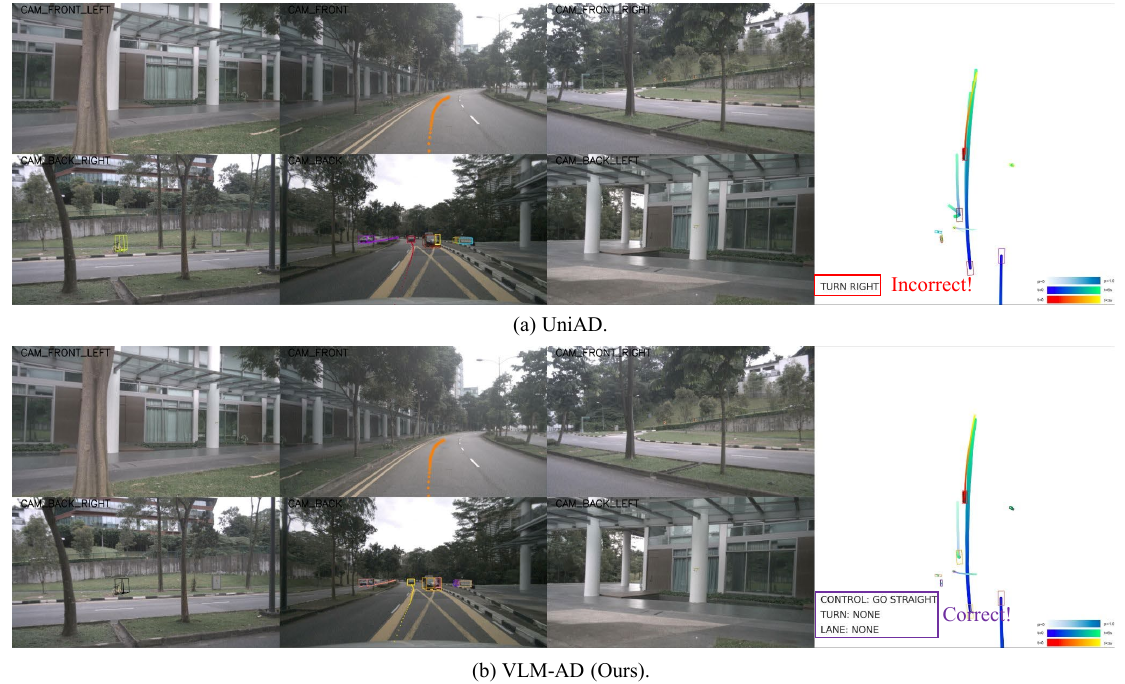}
   \caption{The baseline model predicts an accurate trajectory but is based on an incorrect command, while our method predicts both accurate actions and a precise future trajectory.}
   \label{fig:supp_vis_7}
\end{figure*}

\begin{figure*}[ht]
  \centering
   \includegraphics[width=0.98\linewidth]{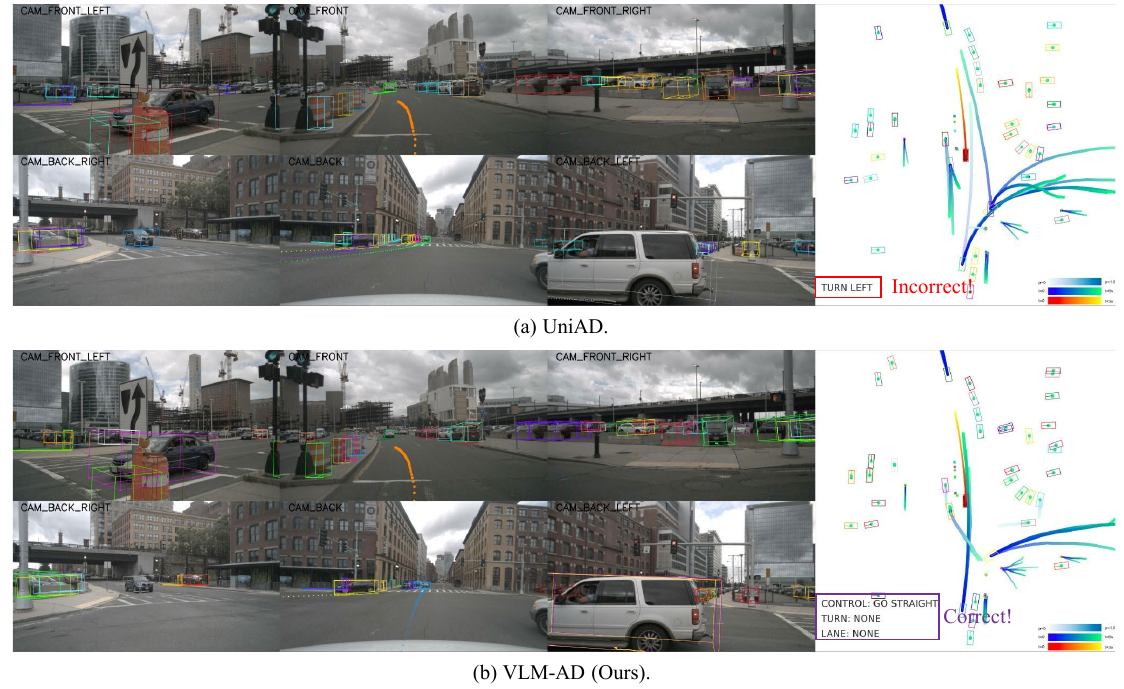}
   \caption{The baseline model predicts an accurate trajectory but is based on an incorrect command, while our method predicts both accurate actions and a precise future trajectory.}
   \label{fig:supp_vis_8}
\end{figure*}

We present a total of eight qualitative examples:
\cref{fig:supp_vis_1}, \cref{fig:supp_vis_2}, \cref{fig:supp_vis_3}, \cref{fig:supp_vis_4}, \cref{fig:supp_vis_5}, \cref{fig:supp_vis_6}, \cref{fig:supp_vis_7}, and \cref{fig:supp_vis_8} to compare the planning results of our proposed method against UniAD.

In \cref{fig:supp_vis_1}, \cref{fig:supp_vis_3}, \cref{fig:supp_vis_4}, \cref{fig:supp_vis_5} and \cref{fig:supp_vis_6}, the planning trajectories generated by UniAD are winding, lack smoothness, and fail to stay within the center of the lane. In contrast, our method produces trajectories that are significantly smoother and stay within the lane boundaries.

Similarly, in \cref{fig:supp_vis_2}, \cref{fig:supp_vis_3}, \cref{fig:supp_vis_4}, \cref{fig:supp_vis_7}, and \cref{fig:supp_vis_8}, the commands generated by UniAD are incorrect, as the ego vehicle is moving forward. However, our action head successfully predicts the correct actions in these scenarios.

These qualitative examples highlight the capability of VLM-AD to produce smoother and more accurate planning trajectories in challenging driving scenarios, while also providing enhanced interpretability.



\end{document}